\vspace{-1em}
\pdfoutput=1

\documentclass[11pt]{article}

\usepackage[]{acl}

\usepackage{times}
\usepackage{latexsym}
\usepackage{geometry}
\usepackage{pdflscape}
\usepackage{tabto}
\definecolor{darkgreen}{rgb}{0.0, 0.5, 0.0}

\usepackage[T1]{fontenc}

\usepackage[utf8]{inputenc}

\usepackage{microtype}

\usepackage{booktabs}
\usepackage{algorithm}
\usepackage{algpseudocode}
\usepackage{amsmath}
\usepackage{amsfonts}
\usepackage{amssymb}
\usepackage{graphicx}
\usepackage{enumitem}
\usepackage{colortbl}
\usepackage{hhline}
\usepackage{booktabs}
\everypar{\looseness=-1}
\DeclareMathOperator*{\argmax}{arg\,max}
\DeclareMathOperator*{\argmin}{arg\,min}
\usepackage{rotating}

\title{Bilingual Lexicon Induction for Low-Resource Languages \\ using
Graph Matching via Optimal Transport} 

\author{Kelly Marchisio\textsuperscript{1}, Ali Saad-Eldin\textsuperscript{3}, Kevin Duh\textsuperscript{1,4}, \\ \bf{Carey Priebe\textsuperscript{2,4}, Philipp Koehn\textsuperscript{1}} \\
        Department of \textsuperscript{1}Computer Science, \textsuperscript{2}Department of Applied Mathematics and Statistics, \\
        \textsuperscript{3}Department of Biomedical Engineering, 
        \textsuperscript{4}Human Language Technology Center of Excellence\\ 
        Johns Hopkins University \\ 
        {\tt \{kmarc,asaadel1\}@jhu.edu} \\
        {\tt kevinduh@cs.jhu.edu, \{cep, phi\}@jhu.edu}}

\date{}

\begin{document}
\maketitle
\begin{abstract}
Bilingual lexicons form a critical component of various natural language processing applications, including unsupervised and semisupervised machine translation and crosslingual information retrieval. We improve bilingual lexicon induction performance across 40 language pairs with a graph-matching method based on optimal transport. The method is especially strong with low amounts of supervision.  

\end{abstract}

\section{Introduction}
Bilingual lexicon induction (BLI) from word embedding spaces is a popular task with a large body of existing literature \cite[e.g.][]{mikolov2013exploiting, artetxe-etal-2018-robust, conneau-lample-2018,patra-etal-2019-bilingual,shi-etal-2021-bilingual}. The goal is to extract a dictionary of translation pairs given separate language-specific embedding spaces, which can then be used to bootstrap downstream tasks such as cross-lingual information retrieval and unsupervised/semi-supervised machine translation.

A great challenge across NLP is maintaining performance in low-resource scenarios. A common criticism of the BLI and low-resource MT literature is that while claims are made about diverse and under-resourced languages, research is often performed on down-sampled corpora of high-resource, highly-related languages on similar domains \cite{artetxe-etal-2020-call}. Such corpora are not good proxies for true low-resource languages owing to data challenges such as dissimilar scripts, domain shift, noise, and lack of sufficient bitext \cite{marchisio-etal-2020-unsupervised}. These differences can lead to dissimilarity between the embedding spaces (decreasing isometry), causing BLI to fail \cite{sogaard-etal-2018-limitations,nakashole-flauger-2018-characterizing,ormazabal-etal-2019-analyzing, glavas-etal-2019-properly, vulic-etal-2019-really, patra-etal-2019-bilingual, marchisio-etal-2020-unsupervised}.

There are two axes by which a language dataset is considered ``low-resource". First, the language itself may be a low-resource language: one for which little bitext and/or monolingual text exists. Even for high-resource languages, the long tail of words may have poorly trained word embeddings due rarity in the dataset \cite{gong2018frage, czarnowska2019don}. In the data-poor setting of true low-resource languages, a great majority of words have little representation in the corpus, resulting in poorly-trained embeddings for a large proportion of them. 
The second axis is low-supervision. Here, there are few ground-truth examples from which to learn. For BLI from word embedding spaces, low-supervision means there are few seeds from which to induce a relationship between spaces, regardless of the quality of the spaces themselves.  

We bring a new algorithm for graph-matching based on optimal transport (OT) to the NLP and BLI literature. We evaluate using 40 language pairs under varying amounts of supervision. The method works strikingly well across language pairs, especially in low-supervision contexts. As low-supervision on low-resource languages reflects the real-world use case for BLI, this is an encouraging development on realistic scenarios. 

\section{Background}

The typical baseline approach for BLI from word embedding spaces assumes that spaces can be mapped via linear transformation. Such methods typically involve solutions to the Procrustes problem (see \citet{gower2004procrustes} for a review). Alternatively, a graph-based view considers words as nodes in undirected weighted graphs, where edges are the distance between words. Methods taking this view do not assume a linear mapping of the spaces exists, allowing for more flexible matching.

\paragraph{BLI from word embedding spaces} Assume separately-trained monolingual word embedding spaces: $\bf{X} \in \mathbb{R}^{n \times d}$, $\bf{Y} \in \mathbb{R}^{m \times d}$ where $n$/$m$ are the source/target language vocabulary sizes and $d$ is the embedding dimension. We build the matrices $\overline{\bf{X}}$ and $\overline{\bf{Y}}$ of seeds from $\bf{X}$ and $\bf{Y}$, respectively, such that given $s$ seed pairs ${(x_1, y_1), (x_2, y_2), ... (x_s, y_s)}$, the first row of $\overline{\bf{X}}$ is $x_1$, the second row is $x_2$, etc.  We build $\overline{\bf{Y}}$ analogously for the $y$-component of each seed pair. The goal is to recover matches for the $\bf{X} \setminus \overline{\bf{X}}$ and/or $\bf{Y} \setminus \overline{\bf{Y}}$ non-seed words. 

\paragraph{Procrustes} Many BLI methods use solutions to the Procrustes problem \cite[e.g.][]{artetxe-etal-2019-effective, conneau-lample-2018, patra-etal-2019-bilingual}. These compute the optimal transform $\bf{W}$ to map seeds: 
\begin{equation}
    \min_{\bf{W}\in\mathbb{R}^{d \times d}} ||\overline{\bf{X}}\bf{W}-\overline{\bf{Y}}||_F^2
\label{eq:proc-eq}
\end{equation}
Once solved for $\bf{W}$, then $\bf{XW}$ and $\bf{Y}$ live in a shared space and translation pairs can be extracted via nearest-neighbor search. Constrained to the space of orthogonal matrices, Equation \ref{eq:proc-eq} has a simple closed-form solution \cite{schonemann1966generalized}:  
\begin{center}
    $\bf{W} = VU^T$ \quad $U\Sigma V = \text{SVD}(\overline{\bf{Y}}^T\overline{\bf{X}})$
\end{center}

\paragraph{Graph View}
Here, words are nodes in monolingual graphs $\bf{G_x}, \bf{G_y} \in \mathbb{R}^{n \times n}$
, and cells in $\bf{G_x}, \bf{G_y}$ are edge weights representing distance between words. As is common in NLP, we use cosine similarity. The objective function is Equation \ref{eq:quad-assign}, where $\Pi$ is the set of \textit{\textbf{permutation matrices}}.\footnote{A permutation matrix represents a one-to-one mapping: There is a single $1$ in each row and column, and $0$ elsewhere.} 
Intuitively, $\bf{P G_y P}^T$ finds the optimal relabeling of $\bf{G_y}$ to align with $\bf{G_x}$. This ``minimizes edge-disagreements" between $\bf{G_x}$ and $\bf{G_y}$. This graph-matching objective is NP-Hard. Equation \ref{eq:quad-assign-tr} is equivalent.
\begin{equation}
\label{eq:quad-assign}
    \min_{\bf{P}\in \Pi} ||\bf{G_x} -\bf{P G_y P}^T||_F^2
\end{equation}
\begin{equation}
\label{eq:quad-assign-tr}
    \max_{\bf{P}\in \Pi} \textit{trace}(\bf{G_x}^TP\bf{G_y}P^T)
\end{equation}
Ex. Take source words $x_1$, $x_2$. We wish to recover valid translations $y_{x_1}$, $y_{x_2}$.  If $\text{distance}(x_1,x_2)\!\!=\!\!\text{distance}(y_{x_1},y_{x_2})$, a solution $P$ can have an edge-disagreement of $0$ here.  We then extract $y_{x_1}, y_{x_2}$ as translations of $x_1$, $x_2$. In reality, though, it is unlikely that $\text{distance}(x_1,x_2)=\text{distance}(y_{x_1},y_{x_2})$.  Because Equation \ref{eq:quad-assign} finds the ideal $P$ to minimize edge disagreements over the entire graphs, we hope that nodes paired by $P$ are valid translations. If $\bf{G_x}$ and $\bf{G_y}$ are isomorphic and there is a unique solution, then $P$ correctly recovers all translations.

Graph-matching is an active research field and is computationally prohibitive on large graphs, but approximation algorithms exist. BLI involves matching large, non-isomorphic graphs---among the greatest challenges for graph-matching.

\subsection{FAQ Algorithm for Graph Matching}
\citet{vogelstein2015fast}'s Fast Approximate Quadratic Assignment Problem algorithm (FAQ) uses gradient ascent to approximate a solution to Equation \ref{eq:quad-assign}. 
Motivated by ``connectonomics" in neuroscience (the study of brain graphs with biological [groups of] neurons as nodes and neuronal connections as edges), FAQ was designed to perform accurately and efficiently on large graphs.

FAQ relaxes the search space of Equation \ref{eq:quad-assign-tr} to allow any doubly-stochastic matrix (the set $\mathcal{D}$). Each cell in a doubly-stochastic matrix is a non-negative real number and each row/column sums to 1. The set $\mathcal{D}$ thus contains $\Pi$ but is much larger. Relaxing the search space makes it easier to optimize Equation \ref{eq:quad-assign-tr} via gradient ascent/descent.\footnote{``descent" for the Quadratic Assignment Problem, ``ascent" for the Graph Matching Problem. The optimization objectives are equivalent: See \citet{vogelstein2015fast} for a proof.}  FAQ solves the objective 
with the Frank-Wolfe method \cite{frank1956algorithm} then projects back to a permutation matrix.

Algorithm \ref{alg:faq} is FAQ; $\mathit{f}(P)$ = $\textit{trace}(\bf{G_x}^TP\bf{G_y}P^T)$. These may be built as $\bf{G_x} = \bf{XX^T}$ and $\bf{G_y} = \bf{YY^T}$.  $\bf{G_x}$ and $\bf{G_y}$ need not have the same dimensionality. 
Step 2 finds a permutation matrix approximation $Q^{\{i\}}$ to $P^{\{i\}}$ in the direction of the gradient.  
Finding such a P 
requires approximation when P is high-dimensional. Here, it is solved via the \textbf{Hungarian Algorithm} \cite{kuhn1955hungarian, jonker1987shortest}, whose solution is a permutation matrix. 
Finally, $P^{n}$ is projected back onto to the space of permutation matrices. 
Seeded Graph Matching \cite[SGM; ][]{fishkind2019seeded} is a variant of FAQ allowing for supervision, and was recently shown to be effective for BLI by \citet{marchisio-etal-2021-analysis-euclidean}. 
\begin{algorithm} \footnotesize
\caption{FAQ Algorithm for Graph Matching}\label{alg:faq}
    \begin{algorithmic}
    \State \textbf{Let:} $\bf{G_x}, \bf{G_y} \in \mathbb{R}^{n \times n}$, $P^{\{0\}} \in \mathcal{D}$ (dbl-stoch.) 
    \While{stopping criterion not met}
    \State 1. Calculate $\nabla f(P^{\{i\}})$: 
    
    \quad$\nabla f(P^{\{i\}})=G_xP^{\{i\}}G_y^T + G_x^TP^{\{i\}}G_y$ 
    \State 2.  $Q^{\{i\}} = $ permutation matrix approx. to $\nabla f(P^{\{i\}})$ 
    
    \quad via Hungarian Algorithm
    \State 3. Calculate step size: 
    
    \quad\quad $\argmax\limits_{\alpha \in [0,1]}\mathit{f}(\alpha P^{\{i\}} + (1-\alpha) Q^{\{i\}})$
    \State 4. Update $P^{\{i+1\}}:= \alpha P^{\{i\}} + (1-\alpha) Q^{\{i\}}$
    \EndWhile
    \State \Return permutation matrix approx. to $P^{\{n\}}$ via Hung. Alg.
    \end{algorithmic}
\end{algorithm}
The interested reader may find \citet{vogelstein2015fast} and \citet{fishkind2019seeded} enlightening for descriptive intuitions of the FAQ and SGM algorithms.\footnote{'Section 3: Fast Approximate QAP Algorithm' \cite{vogelstein2015fast}, 'Section 2.2. From FAQ to SGM' \cite{fishkind2019seeded}.}     

\paragraph{Strengths/Weaknesses}
FAQ/SGM perform well solving the exact graph-matching problem: where graphs are isomorphic and a full matching exists.  In reality, large graphs are rarely isomorphic. For BLI, languages have differing vocabulary size, synonyms/antonyms, and idiosyncratic concepts; it is more natural to assume that an exact matching between word spaces does \textit{not} exist, and that multiple matchings may be equally valid. This is an inexact graph-matching problem. FAQ generally performs poorly finding non-seeded inexact matchings \cite{saad2021graph}.

\subsection{GOAT}
Graph Matching via OptimAl Transport (GOAT) \cite{saad2021graph} is a new graph-matching method which uses advances in OT.  Similar to SGM, GOAT amends FAQ and can use seeds. GOAT has been successful for the inexact graph-matching problem on non-isomorphic graphs: whereas FAQ rapidly fails on non-isomorphic graphs, GOAT maintains strong performance. 

\paragraph{Optimal Transport}
OT is an optimization problem concerned with the most efficient way to transfer probability mass from distribution $\mu$ to distribution $\mathit{v}$. Discrete\footnote{As ours is, as we compute over matrices.} OT minimizes the inner product of a transportation ``plan" matrix $P$ with a cost matrix $C$, as in Equation \ref{eq:ot}. $\langle \cdot , \cdot \rangle$ is the Frobenius inner product.
\begin{equation}
    P^*= \argmin \limits_{P \in \mathcal{U}(r,c)} \langle P, C\rangle
    \label{eq:ot}
\end{equation}
$P$ is an element of the ``transportation polytope" $U(r,c)$---the set of matrices whose rows sum to $r$ and columns sum to $c$. The Hungarian Algorithm approximately solves OT, but the search space is restricted to permutation matrices. 

\paragraph{Sinkhorn: Lightspeed OT} \citet{cuturi2013sinkhorn} introduce Sinkhorn distance, an approximation of OT distance 
that can be solved quickly and accurately by adding an entropy penalty $h$ to Equation \ref{eq:ot}. Adding $h$ makes the objective easier and more efficient to compute, and encourages ``intermediary" solutions similar to that seen in the \textbf{Intuition} subsection. 
\begin{equation}
    P^\lambda = \argmin \limits_{P \in \mathcal{U}(r,c)} \langle P, C\rangle - \frac{1}{\lambda}h(P)
    \label{eq:sinkhorn}
\end{equation}
As $\lambda\rightarrow\infty$, $P^\lambda$ approaches the ideal transportation matrix $P^*$. \citet{cuturi2013sinkhorn} show that Equation \ref{eq:sinkhorn} can be computed using Sinkhorn's 
algorithm \cite{sinkhorn1967diagonal}. 
The interested reader can see details of the algorithm in \citet{cuturi2013sinkhorn,COTFNT}. 
Unlike the Hungarian Algorithm, Sinkhorn has no restriction to a permutation matrix solution and can be solved over any $U(r,c)$.

\paragraph{Sinkhorn in GOAT} GOAT uses \citet{cuturi2013sinkhorn}'s algorithm to solve Equation \ref{eq:sinkhorn} over $U(1,1)$, the set of doubly-stochastic matrices $\mathcal{D}$. They call this the ``doubly stochastic OT problem", and the algorithm that solves it ``Lightspeed Optimal Transport" (LOT).
Although Sinkhorn distance was created for efficiency, \citet{saad2021graph} find that using the matrix $P^\lambda$ that minimizes Sinkhorn distance also improves matching performance on large and non-isometric graphs.  Algorithm \ref{alg:goat} is GOAT. 
\begin{algorithm} \footnotesize
\caption{GOAT}\label{alg:goat}
    \begin{algorithmic}
    \State \textbf{Let:} $\bf{G_x}, \bf{G_y} \in \mathbb{R}^{n \times n}$, $P^{\{0\}} \in \mathcal{D}$ (dbl-stoch.) 
    \While{stopping criterion not met}
    \State 1. Calculate $\nabla f(P^{\{i\}})$: 
    
    \quad$\nabla f(P^{\{i\}})=G_xP^{\{i\}}G_y^T + G_x^TP^{\{i\}}G_y$ 
    \State 2.  $Q^{\{i\}} = $ dbl-stoch. approx. to $\nabla f(P^{\{i\}})$ via LOT.
    \State 3. Calculate step size: 
    
    \quad\quad $\argmax\limits_{\alpha \in [0,1]}\mathit{f}(\alpha P^{\{i\}} + (1-\alpha) Q^{\{i\}})$
    \State 4. Update $P^{\{i+1\}}:= \alpha P^{\{i\}} + (1-\alpha) Q^{\{i\}}$
    \EndWhile
    \State \Return permutation matrix approx. to $P^{\{n\}}$ via Hung. Alg.
    \end{algorithmic}
\end{algorithm}
\paragraph{Intuition} The critical difference between SGM/FAQ and GOAT is how each calculates step direction based on the gradient. Under the hood, each algorithm maximizes $\mathit{trace}(Q^T\nabla f(P^{\{i\}})$ to compute $Q^{\{i\}}$ (the step direction) in Step 2 of their respective algorithms. See \citet{saad2021graph} or \citet{fishkind2019seeded} for a derivation. FAQ uses the Hungarian Algorithm and GOAT uses LOT.

For $\nabla f(P^{\{i\}})$ below, there are \textit{two} valid permutation matrices $Q_1$ and $Q_2$ that maximize the trace. When multiple solutions exist, the Hungarian Algorithm chooses one arbitrarily. Thus, updates of $P$ in FAQ are constrained to be permutation matrices. 
\[
\nabla f(P^{\{i\}}) = \begin{pmatrix}
0 & 3 & 0\\
2 & 1 & 2\\
0 & 0 & 0
\end{pmatrix}
\]
\[
Q_1 = \begin{pmatrix}
0 & 1 & 0\\
0 & 0 & 1\\
1 & 0 & 0
\end{pmatrix}, \quad
Q_2 = \begin{pmatrix}
0 & 1 & 0\\
1 & 0 & 0\\
0 & 0 & 1
\end{pmatrix}
\]
\[
\textit{trace}(Q_1^T\nabla f(P^{\{i\}})) = \textit{trace}(Q_2^T\nabla f(P^{\{i\}})) = 5
\]

Concerningly, \citet{saad2021graph} find that seed order influences the solution in a popular implementation of the Hungarian Algorithm.  Since BLI is a high-dimensional many-to-many task, arbitrary choices could meaningfully affect the result.

\begin{figure}[]
\hfill\includegraphics[height=0.13\textheight,width=1\linewidth]{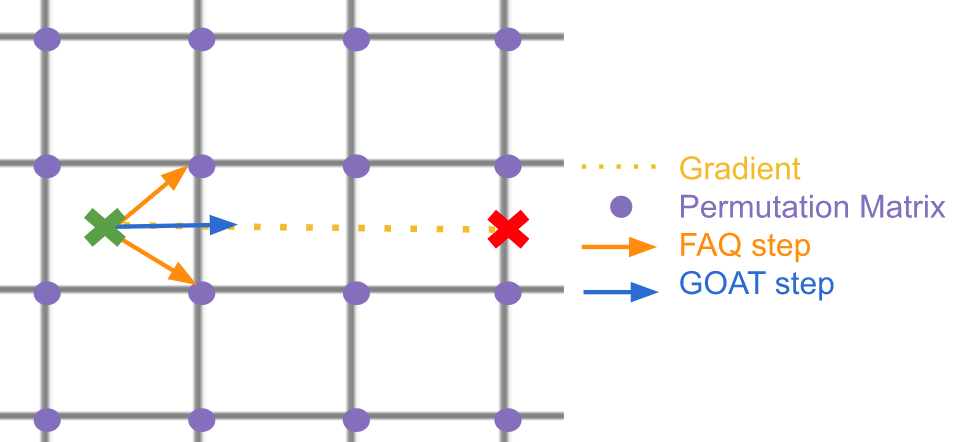}
\centering
\includegraphics[height=0.14\textheight,width=1\linewidth]{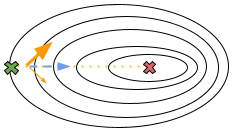}
\caption{Optimization step of FAQ vs. GOAT. FAQ arbitrarily chooses the direction of a permutation matrix. GOAT averages permutation matrices to take a smoother path.}
\label{fig:sgm-v-goat}
\end{figure}

GOAT, on the other hand, can step in the direction of a doubly-stochastic matrix. 
\citet{saad2021graph} prove that given multiple permutation matrices that equally approximate the gradient at $P^{\{i\}}$, any convex linear combination is a doubly stochastic matrix that equally approximates the gradient:
\begin{equation*}
    P_{\lambda} = \sum\limits_i^n \lambda_iP_i \\
     \quad s.t.\;\lambda_1 + ... + \lambda_n = 1;\; \lambda_i \in [0,1]
    \label{eq:p_lambda}
\end{equation*}
$P_{\lambda}$ is a weighted combination of \textit{many} valid solutions---obviating the need to arbitrarily select one for the gradient-based update. LOT's output of a doubly-stochastic matrix in Step 2 is similar to finding a $P_{\lambda}$ in that it needn't discretize to a single permutation matrix. In this way, GOAT can be thought of as taking a step that incorporates many possible permutation solutions. For instance, GOAT may select $Q_{ds} = \frac{1}{2}Q_1 + \frac{1}{2}Q_2$, which also maximizes $\mathit{trace}(Q^T\nabla f(P^{\{i\}})$. 
\begin{equation*}
    Q_{ds} = \begin{pmatrix}
    0 & 1 & 0\\
    \frac{1}{2} & 0 & \frac{1}{2}\\
    \frac{1}{2} & 0 & \frac{1}{2}
    \end{pmatrix}
\end{equation*}
\begin{equation*}
\textit{trace}(Q_{ds}^T\nabla f(P^{\{i\}})) = 5
\end{equation*}
Thus whereas FAQ takes non-deterministic ``choppy" update steps, GOAT optimizes smoothly and deterministically. 
Figure \ref{fig:sgm-v-goat} is an illustration.

\section{Experimental Setup}
We run Procrustes, SGM, and GOAT on 40 language pairs. We also run system combination experiments similar to \citet{marchisio-etal-2021-analysis-euclidean}.  We evaluate with the standard precision@1 (P@1). 

We induce lexicons using \textbf{(1)} the closed-form solution to the orthogonal Procrustes problem of Equation \ref{eq:proc-eq}, extracting nearest neighbors using CSLS \cite{conneau-lample-2018}, \textbf{(2)} SGM, solving the seeded version of Equation \ref{eq:quad-assign}, and \textbf{(3)} GOAT. Word graphs are $\bf{G_x} = \bf{XX^T}$, $\bf{G_y} = \bf{YY^T}$. 

\begin{figure}[]
\centering
\includegraphics[width=1\linewidth]{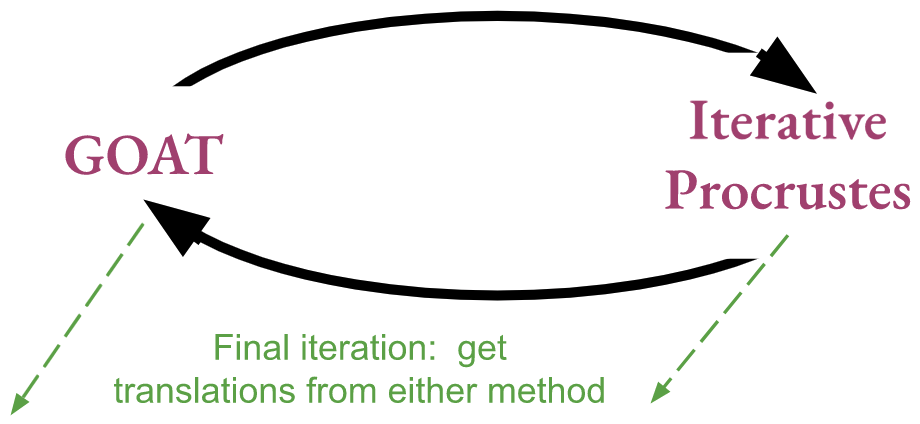}
\caption{Combination system: Iterative Procrustes (IterProc) \& GOAT. (1) run GOAT in forward/reverse directions (2) intersect hypotheses, pass to IterProc, (3) run IterProc forward \& reverse (4) intersect hypotheses, pass to Step (1). Repeat $N$ cycles. (End) Get final translations from forward run of GOAT or IterProc.}
\label{fig:goat_combo}
\end{figure}

\paragraph{System Combination} We perform system combination experiments analogous to those of \citet{marchisio-etal-2021-analysis-euclidean}, incorporating GOAT. Figure \ref{fig:goat_combo} shows the system, which is made of two components: GOAT run in forward and reverse directions, and ``Iterative Procrustes with Stochastic-Add" from \citet{marchisio-etal-2021-analysis-euclidean}. This iterative version of Procrustes runs Procrustes in source$\rightarrow$target and target$\rightarrow$source directions and feeds H random hypotheses from the intersection of both directions into another run of Procrustes with the gold seeds. The process repeats for $I$ iterations, adding $H$ more random hypotheses each time until all are chosen. We set $H = 100$ and $I = 5$, as in the original work. 

\subsection{Data \& Software}
We use publicly-available fastText word embeddings \cite{bojanowski-etal-2017-enriching}\footnote{https://fasttext.cc/docs/en/pretrained-vectors.html} which we normalize, mean-center, and renormalize \cite{artetxe-etal-2018-robust, zhang-etal-2019-girls} and bilingual dictionaries from MUSE\footnote{https://github.com/facebookresearch/MUSE} filtered to be one-to-one.\footnote{For each source word, keep the first unused target word. Targets are in arbitrary order, so this is random sampling.} For languages with 200,000+ embeddings, we use the first 200,000. Dictionary and embeddings space sizes are in Appendix Table \ref{tab:datasizes}. Each language pair has $\sim$4100-4900 translation pairs post-filtering. We choose 0-4000 pairs in frequency order as seeds for experiments, leaving the rest as the test set.\footnote{Ex. En-De with 100 seeds has 4803 test items. With 1000 seeds, the test set contains 3903 items.} For SGM and GOAT, we use the publicly-available implementations from the GOAT repository\footnote{https://github.com/neurodata/goat. Some exps. used SGM from Graspologic \cite[github.com/microsoft/graspologic;][]{chung2019graspy}, but they are mathematically equal.}  
with default hyperparameters (barycenter initialization). We set reg=500 for GOAT. 
For system combination experiments, we amend the code from \citet{marchisio-etal-2021-analysis-euclidean}\footnote{https://github.com/kellymarchisio/euc-v-graph-bli} to incorporate GOAT.

\subsection{Languages}
The languages chosen reflect various language families and writing systems.  The language families represented are:
\begin{itemize}[topsep=2mm,noitemsep,leftmargin=3mm,listparindent=1mm]
    \item \textbf{Austroasiatic}: \textit{Vietnamese}
    \item \textbf{Austronesian}: \textit{Indonesian, Malay}
    \item \textbf{Dravidian}: \textit{Tamil}
    \item \textbf{Japonic}: \textit{Japanese}
    \item \textbf{Sino-Tibetan}: \textit{Chinese}
    \item \textbf{Uralic}: \textit{Estonian}
    \item \textbf{Indo-European}
    \subitem\hspace{-7mm} \textbf{Balto-Slavic}: \textit{Macedonian, Bosnian, Russian}
    \subitem\hspace{-7mm} \textbf{Germanic}: \textit{English, German}
    \subitem\hspace{-7mm} \textbf{Indo-Iranian}: \textit{Bengali, Persian}
    \subitem\hspace{-7mm} \textbf{Romance}: \textit{French, Spanish, Portuguese, Italian}
\end{itemize}
The chosen languages use varying writing systems, including those using or derived from Latin, Cyrillic, Arabic, Tamil, and character-based scripts.

\section{Results}

\begin{figure*}[htb]
\centering
\makebox[\textwidth]{%
\includegraphics[trim=34 0 34 0, clip, height=0.08\textheight, width=0.195\textwidth]{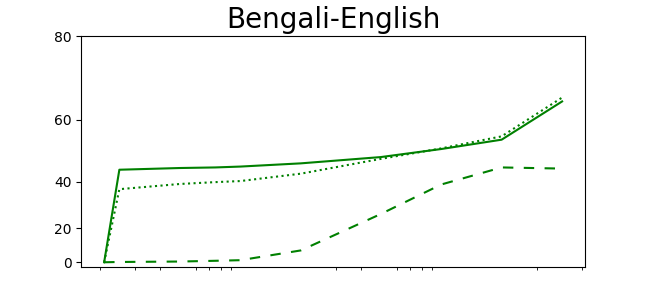}%
\includegraphics[trim=34 0 34 0, clip, height=0.08\textheight, width=0.195\textwidth]{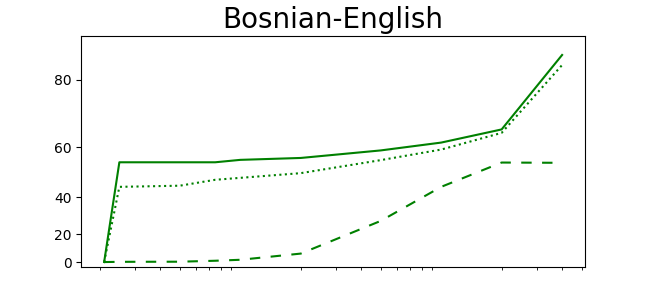}%
\includegraphics[trim=34 0 34 0, clip, height=0.08\textheight, width=0.195\textwidth]{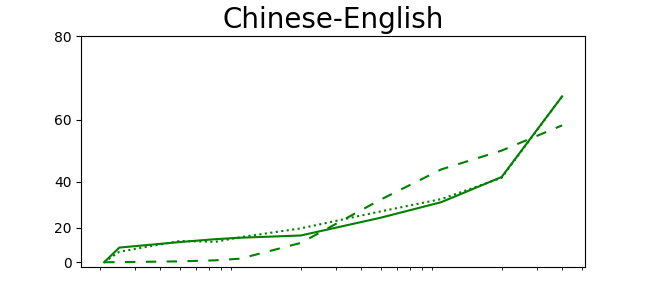}%
\includegraphics[trim=34 0 34 0, clip, height=0.08\textheight, width=0.195\textwidth]{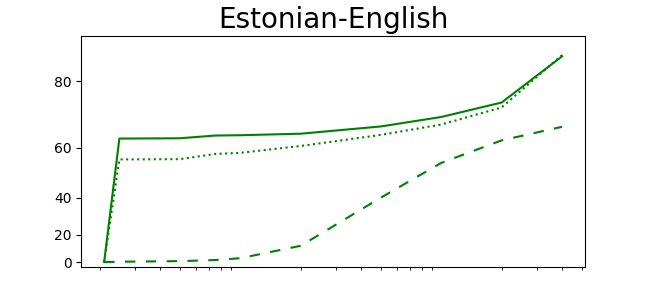}%
\includegraphics[trim=34 0 34 0, clip, height=0.08\textheight, width=0.195\textwidth]{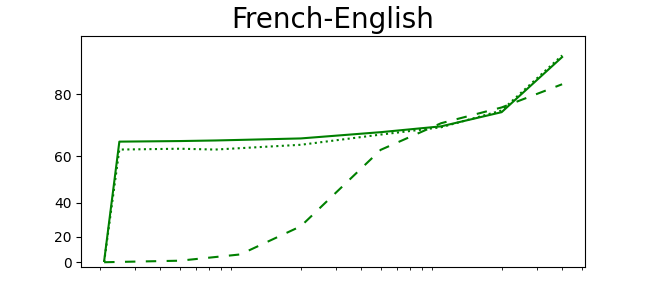}%
}\\
\makebox[\textwidth]{%
\includegraphics[trim=34 0 34 0, clip, height=0.08\textheight, width=0.195\textwidth]{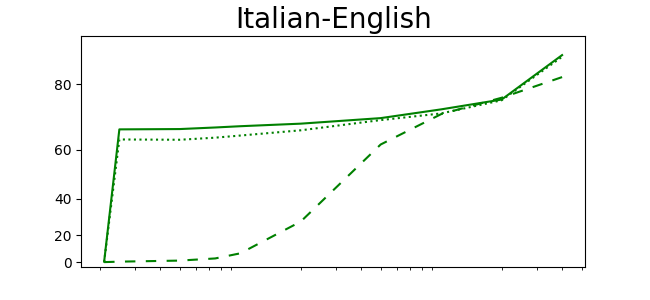}%
\includegraphics[trim=34 0 34 0, clip, height=0.08\textheight, width=0.195\textwidth]{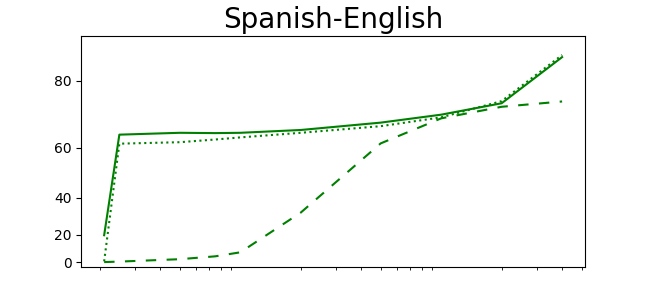}%
\includegraphics[trim=34 0 34 0, clip, height=0.08\textheight, width=0.195\textwidth]{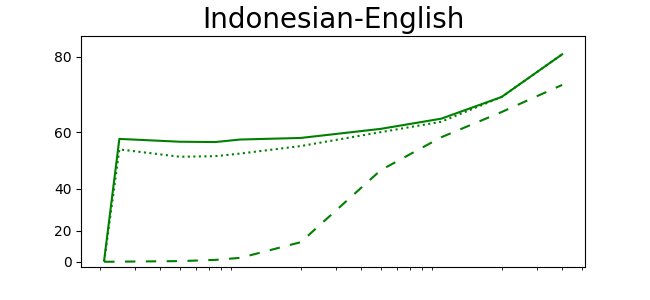}%
\includegraphics[trim=34 0 34 0, clip, height=0.08\textheight, width=0.195\textwidth]{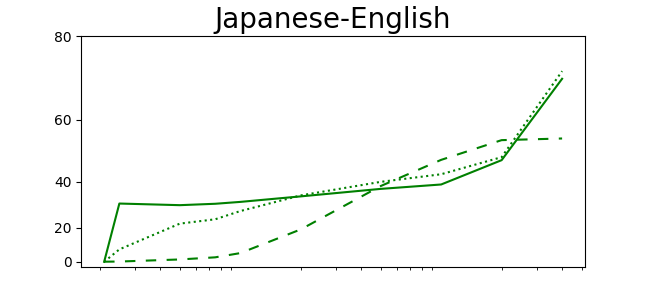}%
\includegraphics[trim=34 0 34 0, clip, height=0.08\textheight, width=0.195\textwidth]{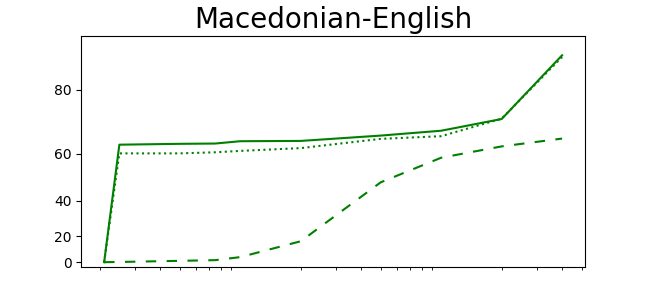}%
}\\
\makebox[\textwidth]{%
\includegraphics[trim=34 0 34 0, clip, height=0.08\textheight, width=0.195\textwidth]{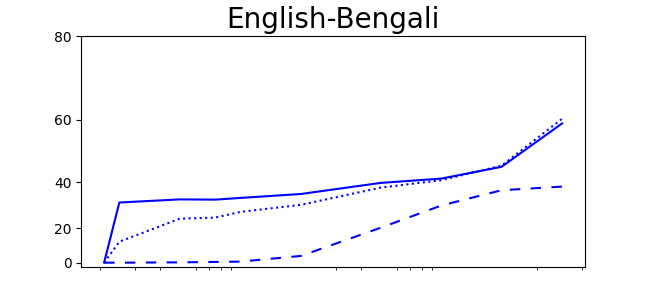}%
\includegraphics[trim=34 0 34 0, clip, height=0.08\textheight, width=0.195\textwidth]{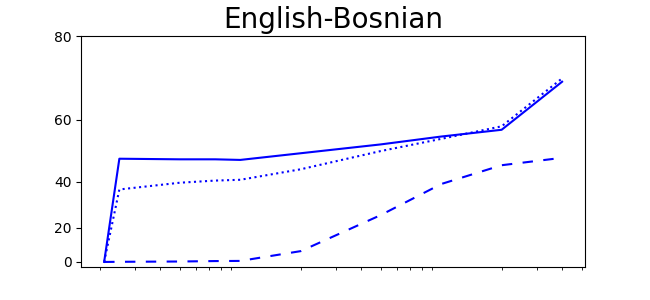}%
\includegraphics[trim=34 0 34 0, clip, height=0.08\textheight, width=0.195\textwidth]{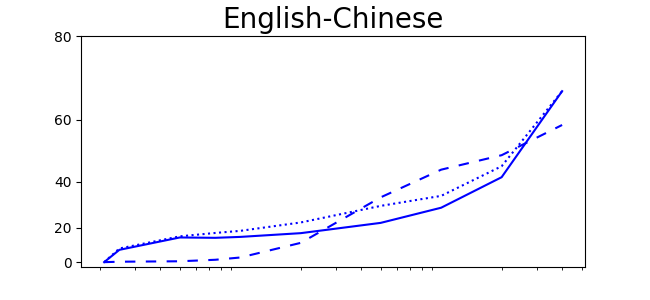}%
\includegraphics[trim=34 0 34 0, clip, height=0.08\textheight, width=0.195\textwidth]{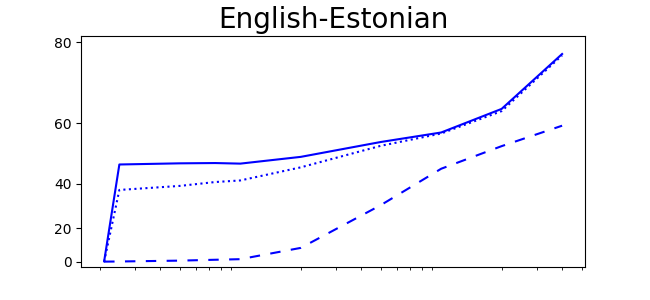}%
\includegraphics[trim=34 0 34 0, clip, height=0.08\textheight, width=0.195\textwidth]{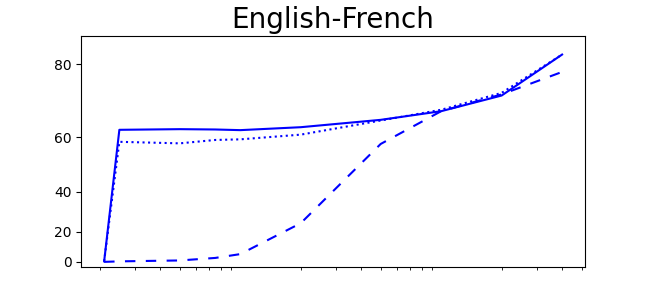}%
}\\
\makebox[\textwidth]{%
\includegraphics[trim=34 0 34 0, clip, height=0.08\textheight, width=0.195\textwidth]{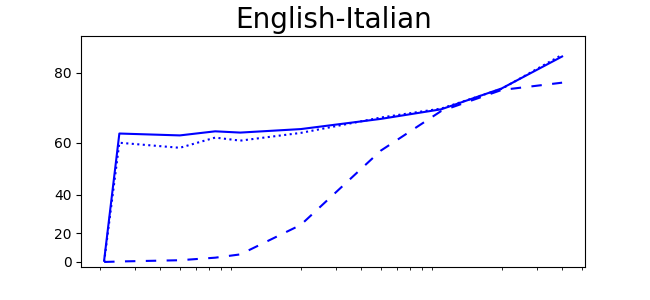}%
\includegraphics[trim=34 0 34 0, clip, height=0.08\textheight, width=0.195\textwidth]{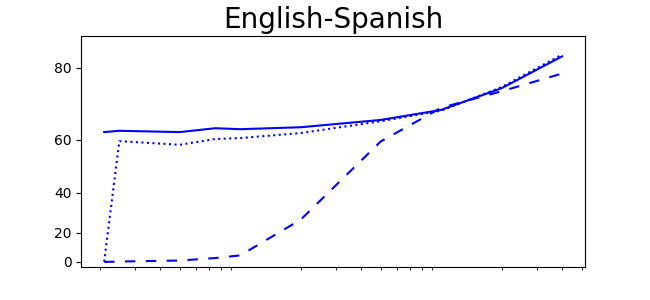}%
\includegraphics[trim=34 0 34 0, clip, height=0.08\textheight, width=0.195\textwidth]{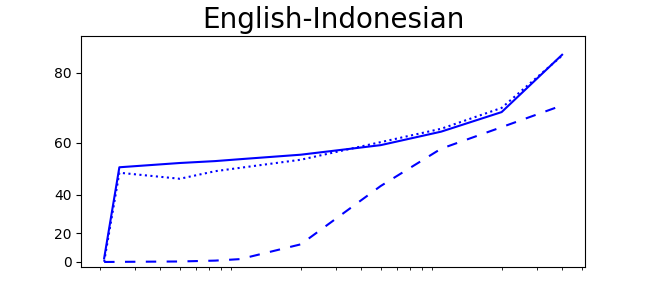}%
\includegraphics[trim=34 0 34 0, clip, height=0.08\textheight, width=0.195\textwidth]{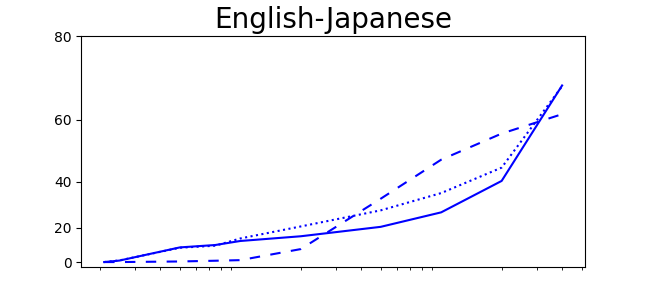}%
\includegraphics[trim=34 0 34 0, clip, height=0.08\textheight, width=0.195\textwidth]{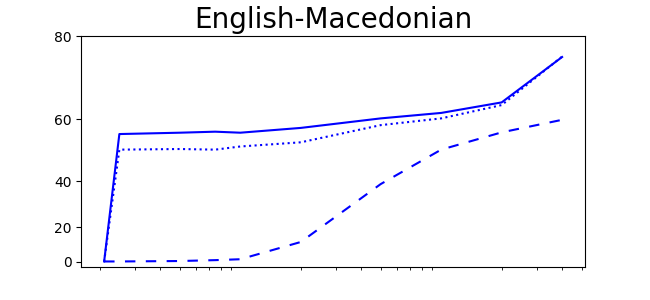}%
}\\
\caption{Visualization of Table \ref{tab:single} for select languages. Procrustes (dashed) vs. SGM (dotted) vs. GOAT (solid). X-axis: \# of seeds (log scale).  Y-axis: Precision@1 ($\uparrow$ is better). GOAT is typically best.}
\label{fig:charts}
\end{figure*}

\begin{table*}[]
    \centering \setlength\tabcolsep{4pt} \footnotesize
    \begin{tabular}{l|rrrc|rrrc|rrrc|rrrc}
    \toprule
    \textit{Seeds}   &\textit{P} &\textit{S} &\textit{G} & $\Delta$ & \textit{P} &\textit{S} &\textit{G} & $\Delta$ &\textit{P} &\textit{S} &\textit{G} & $\Delta$ & \textit{P} &\textit{S} &\textit{G} & $\Delta$  \\
       \midrule
        & \multicolumn{16}{c}{English-to-*} \\
       \midrule
        & \multicolumn{4}{c}{\underline{Bengali}} & \multicolumn{4}{c}{\underline{Bosnian}} & \multicolumn{4}{c}{\underline{Estonian}} & \multicolumn{4}{c}{\underline{Persian}}  \\
25 & 0.0 & 12.9 & \textbf{31.9} & \textcolor{darkgreen}{\textit{(+19.0)}} & 0.1 & 37.0 & \textbf{48.1} & \textcolor{darkgreen}{\textit{(+11.1)}} & 0.1 & 37.5 & \textbf{47.0} & \textcolor{darkgreen}{\textit{(+9.5)}} & 0.2 & 35.2 & \textbf{42.7} & \textcolor{darkgreen}{\textit{(+7.5)}} \\
200 & 4.4 & 30.9 & \textbf{35.4} & \textcolor{darkgreen}{\textit{(+4.5)}} & 7.0 & 44.5 & \textbf{49.9} & \textcolor{darkgreen}{\textit{(+5.4)}} & 9.0 & 46.0 & \textbf{49.6} & \textcolor{darkgreen}{\textit{(+3.6)}} & 7.0 & 40.0 & \textbf{44.4} & \textcolor{darkgreen}{\textit{(+4.4)}} \\
1000 & 30.6 & 40.7 & \textbf{41.3} & \textcolor{darkgreen}{\textit{(+0.6)}} & 39.1 & 54.4 & \textbf{55.1} & \textcolor{darkgreen}{\textit{(+0.7)}} & 45.5 & 57.0 & \textbf{57.3} & \textcolor{darkgreen}{\textit{(+0.3)}} & 40.9 & 48.2 & \textbf{49.9} & \textcolor{darkgreen}{\textit{(+1.7)}} \\
2000 & 36.9 & \textbf{45.9} & 45.5 & \textcolor{red}{\textit{(-0.4)}} & 45.9 & \textbf{58.1} & 57.1 & \textcolor{red}{\textit{(-1.0)}} & 53.1 & 63.4 & \textbf{64.0} & \textcolor{darkgreen}{\textit{(+0.6)}} & 47.6 & 54.1 & \textbf{54.7} & \textcolor{darkgreen}{\textit{(+0.6)}} \\
 & \multicolumn{4}{c}{\underline{Indonesian}}  & \multicolumn{4}{c}{\underline{Macedonian}}  & \multicolumn{4}{c}{\underline{Malay}}  & \multicolumn{4}{c}{\underline{Tamil}}  \\
25 & 0.1 & 49.2 & \textbf{51.3} & \textcolor{darkgreen}{\textit{(+2.1)}} & 0.1 & 51.0 & \textbf{55.8} & \textcolor{darkgreen}{\textit{(+4.8)}} & 0.3 & 24.9 & \textbf{36.6} & \textcolor{darkgreen}{\textit{(+11.7)}} & 0.1 & 1.5 &\textbf{ 1.8} & \textcolor{darkgreen}{\textit{(+0.3)}} \\
200 & 13.0 & 54.1 & \textbf{55.9} & \textcolor{darkgreen}{\textit{(+1.8)}} & 12.2 & 53.3 & \textbf{57.6} & \textcolor{darkgreen}{\textit{(+4.3)}} & 11.5 & 43.3 & \textbf{46.0} & \textcolor{darkgreen}{\textit{(+2.7)}} & 4.1 & 26.7 & \textbf{31.0} & \textcolor{darkgreen}{\textit{(+4.3)}} \\
1000 & 58.0 & \textbf{64.5} & 63.6 & \textcolor{red}{\textit{(-0.9)}} & 51.0 & 60.3 & \textbf{61.8} & \textcolor{darkgreen}{\textit{(+1.5)}} & 48.9 & \textbf{58.9} & 58.3 & \textcolor{red}{\textit{(-0.6)}} & 26.9 & \textbf{36.2} & \textbf{36.2} & \textcolor{darkgreen}{\textit{(+0.0)}} \\
2000 & 65.0 & \textbf{70.7} & 69.5 & \textcolor{red}{\textit{(-1.2)}} & 56.3 & 63.9 & \textbf{64.6} & \textcolor{darkgreen}{\textit{(+0.7)}} & 55.3 & \textbf{65.0} & 63.0 & \textcolor{red}{\textit{(-2.0)}} & 32.3 & \textbf{40.6} & 39.8 & \textcolor{red}{\textit{(-0.8)}} \\
 & \multicolumn{4}{c}{\underline{Vietnamese}}  & \multicolumn{4}{c}{\underline{Chinese}}  & \multicolumn{4}{c}{\underline{Japanese}}  & \multicolumn{4}{c}{\underline{Russian}}  \\
25 & 0.2 & 0.4 &\textbf{ 0.4} & \textcolor{darkgreen}{\textit{(+0.0)}} & 0.3 & \textbf{8.7} & 7.9 & \textcolor{red}{\textit{(-0.8)}} & 0.0 & \textbf{1.1} & 1.0 & \textcolor{red}{\textit{(-0.1)}} & 0.4 & 49.6 & \textbf{55.7} & \textcolor{darkgreen}{\textit{(+6.1)}} \\
200 & 2.9 & 34.8 & \textbf{40.6} & \textcolor{darkgreen}{\textit{(+5.8)}} & 12.0 & \textbf{22.7} & 17.3 & \textcolor{red}{\textit{(-5.4)}} & 8.3 & \textbf{20.7} & 15.6 & \textcolor{red}{\textit{(-5.1)}} & 16.5 & 54.3 & \textbf{57.3} & \textcolor{darkgreen}{\textit{(+3.0)}} \\
1000 & 37.4 & 53.7 & \textbf{54.9} & \textcolor{darkgreen}{\textit{(+1.2)}} & \textbf{44.4} & 34.5 & 29.5 & \textcolor{red}{\textit{(-5.0)}} & \textbf{47.8} & 35.6 & 27.4 & \textcolor{red}{\textit{(-8.2)}} & 58.3 & \textbf{61.7} & 61.5 & \textcolor{red}{\textit{(-0.2)}} \\
2000 & 50.7 & 60.7 & \textbf{61.1} & \textcolor{darkgreen}{\textit{(+0.4)}} & \textbf{49.3} & 45.6 & 41.7 & \textcolor{red}{\textit{(-3.9)}} & \textbf{56.0} & 45.1 & 40.3 & \textcolor{red}{\textit{(-4.8)}} & 65.8 & 67.4 & \textbf{67.5} & \textcolor{darkgreen}{\textit{(+0.1)}} \\
 & \multicolumn{4}{c}{\underline{German}}  & \multicolumn{4}{c}{\underline{French}}  & \multicolumn{4}{c}{\underline{Spanish}}  & \multicolumn{4}{c}{\underline{Italian}}  \\
25 & 0.3 & 44.8 & \textbf{48.5} & \textcolor{darkgreen}{\textit{(+3.7)}} & 0.4 & 58.6 & \textbf{62.4} & \textcolor{darkgreen}{\textit{(+3.8)}} & 0.3 & 59.5 & \textbf{62.8} & \textcolor{darkgreen}{\textit{(+3.3)}} & 0.4 & 60.0 & \textbf{63.0} & \textcolor{darkgreen}{\textit{(+3.0)}} \\
200 & 16.1 & 47.5 & \textbf{50.2} & \textcolor{darkgreen}{\textit{(+2.7)}} & 24.9 & 60.9 & \textbf{63.2} & \textcolor{darkgreen}{\textit{(+2.3)}} & 27.2 & 62.1 & \textbf{63.9} & \textcolor{darkgreen}{\textit{(+1.8)}} & 24.9 & 63.2 & \textbf{64.4} & \textcolor{darkgreen}{\textit{(+1.2)}} \\
1000 & \textbf{57.2} & 54.9 & 54.5 & \textcolor{red}{\textit{(-0.4)}} & 67.9 & \textbf{68.3} & 67.9 & \textcolor{red}{\textit{(-0.4)}} & \textbf{69.6} & 68.8 & 69.0 & \textcolor{darkgreen}{\textit{(+0.2)}} & 69.8 & \textbf{70.5} & 70.3 & \textcolor{red}{\textit{(-0.2)}} \\
2000 & \textbf{63.1} & 61.4 & 61.8 & \textcolor{darkgreen}{\textit{(+0.4)}} & 72.5 & \textbf{72.9} & 72.2 & \textcolor{red}{\textit{(-0.7)}} & 74.1 & \textbf{75.2} & 74.9 & \textcolor{red}{\textit{(-0.3)}} & 75.6 & 75.9 & \textbf{76.0} & \textcolor{darkgreen}{\textit{(+0.1)}} \\
       \midrule
        & \multicolumn{16}{c}{*-to-English} \\
       \midrule
 & \multicolumn{4}{c}{\underline{Bengali}}  & \multicolumn{4}{c}{\underline{Bosnian}}  & \multicolumn{4}{c}{\underline{Estonian}}  & \multicolumn{4}{c}{\underline{Persian}}  \\
25 & 0.2 & 37.2 & \textbf{44.4} & \textcolor{darkgreen}{\textit{(+7.2)}} & 0.2 & 44.7 & \textbf{54.6} & \textcolor{darkgreen}{\textit{(+9.9)}} & 0.3 & 55.9 & \textbf{63.2} & \textcolor{darkgreen}{\textit{(+7.3)}} & 0.1 & 37.1 & \textbf{45.1} & \textcolor{darkgreen}{\textit{(+8.0)}} \\
200 & 7.6 & 43.0 & \textbf{46.6} & \textcolor{darkgreen}{\textit{(+3.6)}} & 6.8 & 50.4 & \textbf{56.2} & \textcolor{darkgreen}{\textit{(+5.8)}} & 12.6 & 60.7 & \textbf{64.8} & \textcolor{darkgreen}{\textit{(+4.1)}} & 9.9 & 42.7 & \textbf{46.8} & \textcolor{darkgreen}{\textit{(+4.1)}} \\
1000 & 39.0 & \textbf{51.5} & 51.3 & \textcolor{red}{\textit{(-0.2)}} & 44.7 & 59.2 & \textbf{61.6} & \textcolor{darkgreen}{\textit{(+2.4)}} & 54.6 & 67.7 & \textbf{70.0} & \textcolor{darkgreen}{\textit{(+2.3)}} & 45.9 & 48.5 & \textbf{49.6} & \textcolor{darkgreen}{\textit{(+1.1)}} \\
2000 & 45.2 & \textbf{55.2} & 54.2 & \textcolor{red}{\textit{(-1.0)}} & 54.5 & 64.8 & \textbf{65.9} & \textcolor{darkgreen}{\textit{(+1.1)}} & 62.6 & 72.8 & \textbf{74.2} & \textcolor{darkgreen}{\textit{(+1.4)}} & 50.3 & \textbf{53.8} & 53.6 & \textcolor{red}{\textit{(-0.2)}} \\
 & \multicolumn{4}{c}{\underline{Indonesian}}  & \multicolumn{4}{c}{\underline{Macedonian}}  & \multicolumn{4}{c}{\underline{Malay}}  & \multicolumn{4}{c}{\underline{Tamil}}  \\
25 & 0.1 & 54.6 & \textbf{58.0} & \textcolor{darkgreen}{\textit{(+3.4)}} & 0.2 & 60.1 & \textbf{63.2} & \textcolor{darkgreen}{\textit{(+3.1)}} & 0.2 & 10.0 & \textbf{38.6} & \textcolor{darkgreen}{\textit{(+28.6)}} & 0.2 & 35.2 & \textbf{44.4} & \textcolor{darkgreen}{\textit{(+9.2)}} \\
200 & 13.3 & 55.7 & \textbf{58.3} & \textcolor{darkgreen}{\textit{(+2.6)}} & 16.5 & 62.0 & \textbf{64.5} & \textcolor{darkgreen}{\textit{(+2.5)}} & 15.5 & 56.2 & \textbf{58.9} & \textcolor{darkgreen}{\textit{(+2.7)}} & 7.4 & 43.4 & \textbf{45.8} & \textcolor{darkgreen}{\textit{(+2.4)}} \\
1000 & 58.5 & 63.2 & \textbf{64.1} & \textcolor{darkgreen}{\textit{(+0.9)}} & 58.5 & 66.1 & \textbf{67.9} & \textcolor{darkgreen}{\textit{(+1.8)}} & 55.7 & 61.6 & \textbf{62.3} & \textcolor{darkgreen}{\textit{(+0.7)}} & 37.1 & 49.6 & \textbf{50.3} & \textcolor{darkgreen}{\textit{(+0.7)}} \\
2000 & 66.0 & \textbf{70.1} & \textbf{70.1} & \textcolor{darkgreen}{\textit{(+0.0)}} & 62.6 & \textbf{71.7} & 71.6 & \textcolor{red}{\textit{(-0.1)}} & 60.2 & \textbf{67.4} & 67.2 & \textcolor{red}{\textit{(-0.2)}} & 45.2 & \textbf{55.1} & 53.5 & \textcolor{red}{\textit{(-1.6)}} \\
 & \multicolumn{4}{c}{\underline{Vietnamese}}  & \multicolumn{4}{c}{\underline{Chinese}}  & \multicolumn{4}{c}{\underline{Japanese}}  & \multicolumn{4}{c}{\underline{Russian}}  \\
25 & 0.1 & 1.0 &\textbf{ 3.3} & \textcolor{darkgreen}{\textit{(+2.3)}} & 0.1 & 6.8 &\textbf{ 9.3} & \textcolor{darkgreen}{\textit{(+2.5)}} & 0.2 & 8.0 & \textbf{31.2} & \textcolor{darkgreen}{\textit{(+23.2)}} & 0.3 & 49.1 & \textbf{54.4} & \textcolor{darkgreen}{\textit{(+5.3)}} \\
200 & 1.7 & 41.3 & \textbf{48.6} & \textcolor{darkgreen}{\textit{(+7.3)}} & 12.0 & \textbf{19.8} & 16.1 & \textcolor{red}{\textit{(-3.7)}} & 19.1 & \textbf{34.6} & 34.2 & \textcolor{red}{\textit{(-0.4)}} & 16.6 & 52.5 & \textbf{55.2} & \textcolor{darkgreen}{\textit{(+2.7)}} \\
1000 & 29.1 & 52.4 & \textbf{57.2} & \textcolor{darkgreen}{\textit{(+4.8)}} & \textbf{44.5} & 33.2 & 31.9 & \textcolor{red}{\textit{(-1.3)}} & \textbf{47.7} & 42.7 & 38.9 & \textcolor{red}{\textit{(-3.8)}} & 56.6 & 58.1 & \textbf{60.3} & \textcolor{darkgreen}{\textit{(+2.2)}} \\
2000 & 56.3 & 64.1 & \textbf{66.3} & \textcolor{darkgreen}{\textit{(+2.2)}} & \textbf{50.8} & 41.5 & 41.8 & \textcolor{darkgreen}{\textit{(+0.3)}} & \textbf{54.0} & 48.6 & 47.6 & \textcolor{red}{\textit{(-1.0)}} & 62.7 & 67.1 & \textbf{68.5} & \textcolor{darkgreen}{\textit{(+1.4)}} \\
 & \multicolumn{4}{c}{\underline{German}}  & \multicolumn{4}{c}{\underline{French}}  & \multicolumn{4}{c}{\underline{Spanish}}  & \multicolumn{4}{c}{\underline{Italian}}  \\
25 & 0.2 & 30.3 & \textbf{34.1} & \textcolor{darkgreen}{\textit{(+3.8)}} & 0.3 & 62.5 & \textbf{65.3} & \textcolor{darkgreen}{\textit{(+2.8)}} & 0.4 & 61.4 & \textbf{64.4} & \textcolor{darkgreen}{\textit{(+3.0)}} & 0.4 & 63.6 & \textbf{66.9} & \textcolor{darkgreen}{\textit{(+3.3)}} \\
200 & 10.6 & 45.0 & \textbf{48.3} & \textcolor{darkgreen}{\textit{(+3.3)}} & 26.8 & 64.2 & \textbf{66.4} & \textcolor{darkgreen}{\textit{(+2.2)}} & 32.6 & 65.0 & \textbf{65.9} & \textcolor{darkgreen}{\textit{(+0.9)}} & 28.2 & 66.6 & \textbf{68.7} & \textcolor{darkgreen}{\textit{(+2.1)}} \\
1000 & 52.8 & 53.1 & \textbf{53.3} & \textcolor{darkgreen}{\textit{(+0.2)}} & \textbf{71.4} & 70.1 & 70.4 & \textcolor{darkgreen}{\textit{(+0.3)}} & 69.5 & 69.8 & \textbf{70.6} & \textcolor{darkgreen}{\textit{(+0.8)}} & 71.6 & 71.8 & \textbf{73.0} & \textcolor{darkgreen}{\textit{(+1.2)}} \\
2000 & \textbf{60.7} & 60.5 & 60.4 & \textcolor{red}{\textit{(-0.1)}} &\textbf{76.2} & 75.3 & 74.8 & \textcolor{red}{\textit{(-0.5)}} & 72.9 & \textbf{74.5} & 73.9 & \textcolor{red}{\textit{(-0.6)}} & \textbf{76.3} & 75.6 & 75.8 & \textcolor{darkgreen}{\textit{(+0.2)}} \\
       \midrule
        & \multicolumn{16}{c}{*-to-*} \\
       \midrule
& \multicolumn{4}{c}{\underline{German-Spanish}}   & \multicolumn{4}{c}{\underline{Italian-French}}   & \multicolumn{4}{c}{\underline{Spanish-Portuguese}}   & \multicolumn{4}{c}{\underline{Portuguese-German}}   \\
25 & 0.5 & 67.4 & \textbf{70.5} & \textcolor{darkgreen}{\textit{(+3.1)}} & 1.1 & 85.4 & \textbf{87.3} & \textcolor{darkgreen}{\textit{(+1.9)}} & 2.5 & 94.3 & \textbf{94.9} & \textcolor{darkgreen}{\textit{(+0.6)}} & 0.5 & 72.5 & \textbf{76.2} & \textcolor{darkgreen}{\textit{(+3.7)}} \\
200  & 26.0 & 67.9 & \textbf{71.0} & \textcolor{darkgreen}{\textit{(+3.1)}} & 57.3 & 86.7 & \textbf{87.7} & \textcolor{darkgreen}{\textit{(+1.0)}} & 74.1 & 94.9 & \textbf{95.2} & \textcolor{darkgreen}{\textit{(+0.3)}} & 36.3 & 74.8 & \textbf{76.9} & \textcolor{darkgreen}{\textit{(+2.1)}} \\
1000 & 71.1 & 74.0 & \textbf{75.4} & \textcolor{darkgreen}{\textit{(+1.4)}} & 86.1 & 89.1 & \textbf{89.4} & \textcolor{darkgreen}{\textit{(+0.3)}} & 93.3 & 95.8 & \textbf{96.0} & \textcolor{darkgreen}{\textit{(+0.2)}} & 75.5 & 78.9 & \textbf{79.6} & \textcolor{darkgreen}{\textit{(+0.7)}} \\
2000 & 76.1 & 78.8 & \textbf{79.1} & \textcolor{darkgreen}{\textit{(+0.3)}} & 88.8 & 89.6 & \textbf{90.4} & \textcolor{darkgreen}{\textit{(+0.8)}} & 94.5 & 97 & \textbf{97.1} & \textcolor{darkgreen}{\textit{(+0.1)}} & 80 & 81.9 & \textbf{82.0} & \textcolor{darkgreen}{\textit{(+0.1)}} \\
& \multicolumn{4}{c}{\underline{Spanish-German}}   & \multicolumn{4}{c}{\underline{French-Italian}}   & \multicolumn{4}{c}{\underline{Portuguese-Spanish}}   & \multicolumn{4}{c}{\underline{German-Portuguese}}   \\
25 & 0.5 & 62.1 & \textbf{66.2} & \textcolor{darkgreen}{\textit{(+4.1)}} & 0.6 & 88.9 & \textbf{90.2} & \textcolor{darkgreen}{\textit{(+1.3)}} & 1.1 & 89.7 & \textbf{90.5} & \textcolor{darkgreen}{\textit{(+0.8)}} & 0.3 & 72.9 & \textbf{76.2} & \textcolor{darkgreen}{\textit{(+3.3)}} \\
200 & 27.7 & 65.2 & \textbf{67.2} & \textcolor{darkgreen}{\textit{(+2.0)}} & 58.4 & 90.4 & \textbf{91.2} & \textcolor{darkgreen}{\textit{(+0.8)}} & 70.0 & 90.4 & \textbf{90.7} & \textcolor{darkgreen}{\textit{(+0.3)}} & 24.3 & 73.9 & \textbf{77.1} & \textcolor{darkgreen}{\textit{(+3.2)}} \\
1000 & 68.8 & 70.6 & \textbf{70.6} & \textcolor{darkgreen}{\textit{(+0.0)}} & 89.6 & 92.1 & \textbf{92.7} & \textcolor{darkgreen}{\textit{(+0.6)}} & 89.5 & 90.4 & \textbf{91.1} & \textcolor{darkgreen}{\textit{(+0.7)}} & 73.7 & 79.8 & \textbf{80.8} & \textcolor{darkgreen}{\textit{(+1.0)}} \\
2000 & 73.6 & 74.0 & \textbf{74.0} & \textcolor{darkgreen}{\textit{(+0.0)}} & 91.8 & 93.9 & \textbf{94.1} & \textcolor{darkgreen}{\textit{(+0.2)}} & 90.9 & 91.7 & \textbf{92.3} & \textcolor{darkgreen}{\textit{(+0.6)}} & 78.7 & 80.6 & \textbf{82.2} & \textcolor{darkgreen}{\textit{(+1.6)}} \\
       \midrule
        & \multicolumn{16}{c}{Averages} \\
       \midrule
 & \multicolumn{4}{c}{\underline{English-to-*}}   & \multicolumn{4}{c}{\underline{*-to-English}}   & \multicolumn{4}{c}{\underline{Non-English}} & \multicolumn{4}{c}{\underline{Overall}}  \\
25 & 0.2 & 33.2 & \textbf{38.6} & \textcolor{darkgreen}{\textit{(+5.4)}} & 0.2 & 38.6 & \textbf{46.3} & \textcolor{darkgreen}{\textit{(+7.7)}} & 0.9 & 79.2 & \textbf{81.5} & \textcolor{darkgreen}{\textit{(+2.3)}} & 0.3 & 37.1 & \textbf{41.9} & \textcolor{darkgreen}{\textit{(+4.8)}} \\
200 & 12.6 & 44.1 & \textbf{46.4} & \textcolor{darkgreen}{\textit{(+2.3)}} & 14.8 & 50.2 & \textbf{52.8} & \textcolor{darkgreen}{\textit{(+2.6)}} & 23.4 & 40.3 & \textbf{41.1} & \textcolor{darkgreen}{\textit{(+0.8)}} & 16.9 & 44.8 & \textbf{46.8} & \textcolor{darkgreen}{\textit{(+2.0)}} \\
1000 & 49.6 & \textbf{54.3} & 53.7 & \textcolor{red}{\textit{(-0.6)}} & 52.3 & 57.4 & \textbf{58.3} & \textcolor{darkgreen}{\textit{(+0.9)}} & 40.5 & 41.9 & \textbf{42.2} & \textcolor{darkgreen}{\textit{(+0.3)}} & 47.5 & 51.2 & \textbf{51.4} & \textcolor{darkgreen}{\textit{(+0.2)}} \\
2000 & 56.2 & \textbf{60.4} & 59.6 & \textcolor{red}{\textit{(-0.8)}} & 59.8 & 63.6 & \textbf{63.7} & \textcolor{darkgreen}{\textit{(+0.1)}} & 42.1 & 43.0 & \textbf{43.2} & \textcolor{darkgreen}{\textit{(+0.2)}} & 52.7 & \textbf{55.7} & 55.5 & \textcolor{red}{\textit{(-0.2)}} \\

        \bottomrule
    \end{tabular}
\caption{P@1 of Procrustes (P), SGM (S) or GOAT (G). $\Delta$ is gain/loss of GOAT vs. SGM. Full results in Appendix. Figure \ref{fig:charts} is a visualization of these results.}
\label{tab:single}
\end{table*}

Results of Procrustes vs. SGM vs. GOAT are in Table \ref{tab:single}, visualized in Figure \ref{fig:charts}. 

\paragraph{Procrustes vs. SGM} \citet{marchisio-etal-2021-analysis-euclidean} conclude that SGM strongly outperforms Procrustes for English$\to$German and Russian$\to$English with 100+ seeds. We find that the trend holds across language pairs, with the effect even stronger with less supervision. SGM performs reasonably with only 50 seeds for nearly all languages, and with only 25 seeds in many. Chinese$\leftrightarrow$English and Japanese$\leftrightarrow$English perform relatively worse, and highly-related languages perform best: French, Spanish, Italian, and Portuguese. German$\leftrightarrow$English performance is low relative to some less-related languages, which have surprisingly strong performance from SGM: Indonesian$\leftrightarrow$English and Macedonian$\leftrightarrow$English score $P@1\! \approx\!50\text{-}60$, even with low supervision. Except for the aforementioned highly-related language pairs, Procrustes does not perform above ${\sim\!10}$ for any language pair with $\leq 100$ seeds, whereas SGM exceeds $P@1 = 10$ with only 25 seeds for 33 of 40 pairs.  

\paragraph{SGM vs. GOAT}
GOAT improves considerably over SGM for nearly all language pairs, and the effect is particularly strong with very low amounts of seeds and less-related languages. GOAT improves upon SGM by +19.0, +8.5, and +7.9 on English$\rightarrow$Bengali with 25, 50, and 75 seeds, respectively. As the major use case of low-resource BLI and MT is dissimilar languages with low supervision, this is an encouraging result. It generally takes 200+ seeds for SGM to achieve similar scores to GOAT with just 25 seeds. 

\subsection{Isomorphism of Embedding Spaces}
\begin{table}
\centering 
\begin{tabular}{c|rr||c|rr}
\toprule
&  \textbf{EVS}  &  \textbf{GH}  &  &  \textbf{EVS}  &  \textbf{GH} \\
\midrule
bn  &  37.79  &  0.49  &  it  &  22.42  &  0.20 \\
bs  &  35.93  &  0.41  &  ja  &  894.20  &  0.55 \\
de  &  11.49  &  0.31  &  mk  &  151.02  &  0.19 \\
es  &  9.91  &  0.21  &  ms  &  153.42  &  0.49 \\
et  &  35.22  &  0.68  &  ru  &  14.19  &  0.46 \\
fa  &  86.98  &  0.39  &  ta  &  56.66  &  0.26 \\
fr  &  27.92  &  0.17  &  vi  &  256.28  &  0.42 \\
id  &  188.98  &  0.39  &  zh  &  519.82  &  0.61 \\
\bottomrule
\end{tabular}
\caption{Degree of isomorphism of embedding spaces in relation to English. EVS=Eigenvector Similarity. GH=Gromov-Hausdorff Distance. $\downarrow$:more isomorphic.}
\label{tab:iso}
\end{table}
Eigenvector similarity \cite[EVS;][]{sogaard-etal-2018-limitations} measures isomorphism of embedding spaces based on the difference of Laplacian eigenvalues. Gromov-Hausdorff distance (GH) measures distance based on nearest neighbors after an optimal orthogonal transformation \cite{patra-etal-2019-bilingual}. EVS and GH are symmetric, and lower means more isometric spaces.  Refer to the original papers for mathematical descriptions.  We compute the metrics over the word embedding using scripts from \citet{vulic-etal-2020-good}\footnote{https://github.com/cambridgeltl/iso-study/scripts} and show results in Table \ref{tab:iso}. We observe a moderate correlation between EVS and GH (Spearman's $\rho=0.434$, Pearson's $r=0.44$). 

\begin{figure}
\centering
\includegraphics[width=0.98\columnwidth]{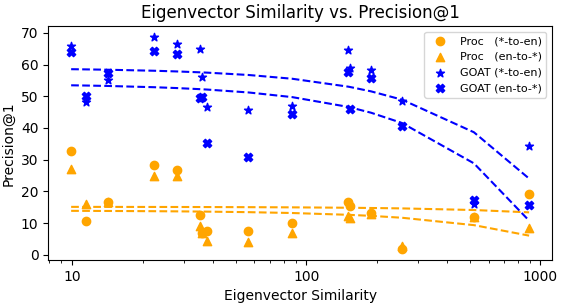}
\includegraphics[width=0.98\columnwidth]{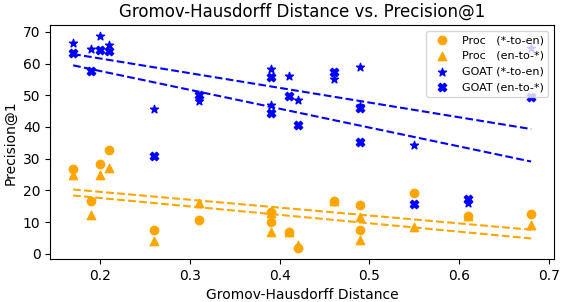}
\caption{X-axis: Eigenvector Similarity (EVS) / Gromov-Hausdorff (GH) Distance of language compared to English. Y-axis: Precision@1 from Procrustes \& GOAT with 200 seeds. $\downarrow$ EVS/GH = $\uparrow$ isomorphic.} 
\label{fig:iso}
\end{figure}

Figure \ref{fig:iso} shows the relationship between relative isomorphism of each language vs. English, and performance of Procrustes/GOAT at 200 seeds. Trends indicate that higher isomorphism varies with higher precision from Procrustes and GOAT. 
GH shows a moderate to strong negative Pearson's correlation with performance from Procrustes and GOAT: $r=-0.47$ and $r=-0.53$, respectively, for *-to-en and -0.55 and -0.61 for en-to-*.  EVS correlates weakly negatively with performance from Procrustes (*-to-en: -0.06, en-to-*: -0.28) and strongly negatively with GOAT (*-to-en: -0.67, en-to-*: ${-0.75}$). As higher GH/EVS indicates less isomorphism, negative correlations imply that lower degrees of isomorphism correlate with lower scores from Procrustes/GOAT. 

\subsection{System Combination}
System combination results are in Table \ref{tab:combined-short}. Similar to \citet{marchisio-etal-2021-analysis-euclidean}'s findings for their combined Procrustes/SGM system, we find (1) our combined Procrustes/GOAT system outperforms Procrustes and GOAT alone, (2) ending with the Iterative Procrustes is best for moderate amounts of seeds, (3) ending with GOAT is best for very low or very high number of seeds. 

Whether we end with Iterative Procrustes vs. GOAT is critically important for the lowest seed sizes: -EndGOAT (-EG) usually fails with 25 seeds; all language pairs except German$\leftrightarrow$English and Russian$\leftrightarrow$English score $P@1\!<\!15.0$, and most score $P@1\!<\!2.0$. Simply switching the order of processing in the combination system, however, boosts performance dramatically: ex. from 0.6 for StartProc-EndGOAT to 61.5 for StartGOAT-EndProc for Bosnian$\to$English with 25 seeds. 

There are some language pairs such as English$\to$Persian and Russian$\leftrightarrow$English where a previous experiment with no seeds had reasonable performance, but the combined system failed.  It is worth investigating where this discrepancy arises.

\begin{table*}[]
    \centering \setlength\tabcolsep{3.5pt} \footnotesize
    \begin{tabular}{l|rrr|rrr|rrr|rrr|rrr|rrr}
    \toprule
        &\textit{Prev} &\textit{-EP} &\textit{-EG} & \textit{Prev} &\textit{-EP} &\textit{-EG} &\textit{Prev} &\textit{-EP} &\textit{-EG} & \textit{Prev} &\textit{-EP} &\textit{-EG} &\textit{Prev} &\textit{-EP} &\textit{-EG} & \textit{Prev} &\textit{-EP} &\textit{-EG} \\
        \hline
        \textit{Seeds} &\multicolumn{3}{c}{\underline{en-bn}} &  \multicolumn{3}{c}{\underline{bn-en}} & \multicolumn{3}{c}{\underline{en-bs}} &  \multicolumn{3}{c}{\underline{bs-en}} &\multicolumn{3}{c}{\underline{en-de}} &  \multicolumn{3}{c}{\underline{de-en}}  \\
25 & \textit{31.9} & \textbf{44.3} & 0.5 & \textit{44.4} & \textbf{53.1} & 6.2 & \textit{48.1} & \textbf{58.4} & 0.4 & \textit{54.6} & \textbf{61.5} & 0.6 & \textit{48.5} & \textbf{61.7} & 59.1 & \textit{34.1} & \textbf{59.3} & 56.8  \\ 
75 & \textit{33.1} & \textbf{44.7} & 39.5 & \textit{45.2} & \textbf{53.5} & 49.1 & \textit{47.9} & \textbf{58.1} & 55.1 & \textit{54.6} & \textbf{61.7} & 57.2 & \textit{48.8} & \textbf{62.3} & 59.7 & \textit{47.0} & \textbf{59.4} & 57.1  \\ 
100 & \textit{33.8} & \textbf{45.3} & 39.7 & \textit{45.5} & \textbf{53.9} & 48.1 & \textit{47.7} & \textbf{58.1} & 55.4 & \textit{55.5} & \textbf{61.3} & 57.5 & \textit{49.0} & \textbf{62.2} & 59.7 & \textit{47.6} & \textbf{59.4} & 56.8  \\ 
2000 & \textit{45.9} & 48.7 & \textbf{49.3} & \textit{55.2} & \textbf{56.6} & 56.3 & \textit{58.1} & 59.9 & \textbf{60.5} & \textit{65.9} & 66.6 & \textbf{69.1} & \textit{63.1} & 66.3 & \textbf{69.4} & \textit{60.7} & 65.1 & \textbf{67.9}  \\ 
4000 & \textit{60.3} & 50.5 & \textbf{61.2} & \textit{65.9} & 55.2 & \textbf{68.9} & \textit{70.6} & 63.4 & \textbf{71.8} & \textbf{\textit{86.1}} & 69.7 & 85.7 & \textit{74.2} & 72.9 & \textbf{79.5} & \textit{71.4} & 67.2 & \textbf{77.6}  \\ 
        \hline
        &\multicolumn{3}{c}{\underline{en-et}} &  \multicolumn{3}{c}{\underline{et-en}} &\multicolumn{3}{c}{\underline{en-fa}} &  \multicolumn{3}{c}{\underline{fa-en}} &\multicolumn{3}{c}{\underline{en-id}} &  \multicolumn{3}{c}{\underline{id-en}}  \\
25 & \textit{47.0} & \textbf{60.4} & 5.9 & \textit{63.2} & \textbf{70.2} & 15.0 & \textit{42.7} & \textbf{54.4} & 4.5 & \textit{45.1} & \textbf{55.1} & 2.0 & \textit{51.3} & \textbf{65.8} & 0.6 & \textit{58.0} & \textbf{66.7} & 1.8  \\
75 & \textit{47.5} & \textbf{60.6} & 58.6 & \textit{64.2} & \textbf{69.8} & 66.5 & \textit{42.9} & \textbf{54.1} & 51.9 & \textit{45.8} & \textbf{55.3} & 52.0 & \textit{53.6} & \textbf{66.0} & 63.3 & \textit{57.0} & \textbf{66.7} & 64.2  \\
100 & \textit{47.3} & \textbf{61.1} & 59.0 & \textit{64.3} & \textbf{70.2} & 66.7 & \textit{43.0} & \textbf{54.3} & 52.3 & \textit{45.6} & \textbf{55.5} & 52.7 & \textit{54.3} & \textbf{66.2} & 63.2 & \textit{57.8} & \textbf{67.1} & 63.9  \\
2000 & \textit{64.0} & 66.6 & \textbf{67.4} & \textit{74.2} & 74.1 & \textbf{75.0} & \textit{54.7} & 58.0 & \textbf{58.4} & \textit{53.8} & \textbf{57.5} & 56.9 & \textit{70.7} & 72.1 & \textbf{74.2} & \textit{70.1} & 72.5 & \textbf{72.9}  \\
4000 & \textit{77.4} & 71.7 & \textbf{80.2} & \textit{86.4} & 80.7 & \textbf{87.2} & \textit{65.9} & 62.4 & \textbf{67.4} & \textit{65.5} & 60.1 & \textbf{67.0} & \textit{84.3} & 76.8 & \textbf{86.2} & \textit{80.5} & 78.2 & \textbf{83.7}  \\
        \hline
        &\multicolumn{3}{c}{\underline{en-mk}} &  \multicolumn{3}{c}{\underline{mk-en}} & \multicolumn{3}{c}{\underline{en-ms}} &  \multicolumn{3}{c}{\underline{ms-en}} &\multicolumn{3}{c}{\underline{en-ru}} &  \multicolumn{3}{c}{\underline{ru-en}}  \\
25 & \textit{55.8} & \textbf{63.9} & 8.0 & \textit{63.2} & \textbf{68.8} & 0.6 & \textit{36.6} & \textbf{62.6} & 0.9 & \textit{38.6} & \textbf{65.3} & 0.2 & \textit{55.7} & \textbf{67.7} & 66.1 & \textit{54.4} & \textbf{63.9} & 62.0  \\
75 & \textit{56.5} & \textbf{64.3} & 63.3 & \textit{63.6} & \textbf{68.8} & 67.9 & \textit{42.6} & \textbf{62.6} & 59.8 & \textit{56.8} & \textbf{65.6} & 62.8 & \textit{55.7} & \textbf{68.1} & 67.0 & \textit{54.8} & \textbf{63.9} & 61.6  \\
100 & \textit{56.2} & 64.1 & \textbf{64.2} & \textit{64.4} & \textbf{69.4} & 67.5 & \textit{42.9} & \textbf{63.0} & 58.6 & \textit{57.7} & \textbf{65.8} & 63.1 & \textit{55.9} & \textbf{67.9} & 66.4 & \textit{55.1} & \textbf{63.8} & 61.3  \\
2000 & \textit{64.6} & 66.8 & \textbf{67.9} & \textit{71.7} & 71.0 & \textbf{73.1} & \textit{65.0} & 67.0 & \textbf{68.5} & \textit{67.4} & 69.4 & \textbf{69.7} & \textit{67.5} & 72.6 & \textbf{74.2} & \textit{68.5} & 69.3 & \textbf{72.5}  \\
4000 & \textit{75.6} & 68.9 & \textbf{77.1} & \textit{88.8} & 74.1 & \textbf{91.1} & \textit{79.3} & 70.7 & \textbf{79.5} & \textit{77.4} & 70.0 & \textbf{79.1} & \textit{83.3} & 79.3 & \textbf{86.5} & \textbf{\textit{89.3}} & 77.4 & \textbf{89.3}   \\        
        \hline
        &\multicolumn{3}{c}{\underline{en-ta}} &  \multicolumn{3}{c}{\underline{ta-en}} &\multicolumn{3}{c}{\underline{en-vi}} &  \multicolumn{3}{c}{\underline{vi-en}} & \multicolumn{3}{c}{\underline{en-zh}} &  \multicolumn{3}{c}{\underline{zh-en}}  \\
25 & \textit{1.8} & \textbf{2.2} & 0.6 & \textit{44.4} & \textbf{51.4} & 2.4 & \textit{0.4} & 0.4 & 0.2 & \textit{3.3} & \textbf{5.3} & 0.2 & \textit{8.7} & \textbf{52.7} & 1.7 & \textit{9.3} & \textbf{48.1} & 0.8  \\
75 & \textit{30.5} & \textbf{40.4} & 35.7 & \textit{45.1} & \textbf{52.4} & 46.9 & \textit{22.9} & \textbf{55.0} & 1.2 & \textit{45.2} & \textbf{59.6} & 54.4 & \textit{17.4} & \textbf{51.6} & 46.4 & \textit{14.1} & \textbf{51.1} & 48.0  \\
100 & \textit{30.6} & \textbf{40.2} & 36.8 & \textit{45.4} & \textbf{52.5} & 47.9 & \textit{30.2} & \textbf{55.6} & 36.2 & \textit{47.1} & \textbf{59.3} & 56.3 & \textit{18.5} & \textbf{51.6} & 47.3 & \textit{15.2} & \textbf{51.0} & 48.0  \\
2000 & \textit{40.6} & 42.7 & \textbf{44.2} & \textit{55.1} & 55.3 & \textbf{56.2} & \textit{61.1} & \textbf{67.3} & 65.8 & \textit{66.3} & \textbf{73.5} & 71.5 & \textit{49.3} & \textbf{58.0} & 57.7 & \textit{41.8} & \textbf{57.6} & 56.8  \\
4000 & \textit{49.1} & 44.0 & \textbf{51.7} & \textbf{\textit{73.8}} & 65.3 & 71.6 & \textit{77.9} & 73.0 & \textbf{80.5} & \textit{82.1} & 80.1 & \textbf{84.3} & \textit{67.5} & 66.4 & \textbf{75.1} & \textit{66.2} & 65.3 & \textbf{73.3} \\
        \bottomrule
    \end{tabular}
    \caption{P@1 of Combination Exps. -EP starts with GOAT, ends with IterProc. -EG: IterProc, ends with GOAT. \textit{Prev} is previous best of prior experiments. Some seed sizes omitted for brevity (see Appendix).}
    \label{tab:combined-short}
\end{table*}

\section{Discussion}
We have seen GOAT's strength in low-resource scenarios and in non-isomorphic embedding spaces.  
As the major use case of low-resource BLI and MT is dissimilar languages with low supervision, GOAT's strong performance is an encouraging result for real-world applications. Furthermore, GOAT outperforms SGM. As the graph-matching objective is NP-hard so all algorithms are approximate, GOAT does a better job by making a better calculation of step direction.  Chinese$\leftrightarrow$English and Japanese$\leftrightarrow$English are outliers, which is worthy of future investigation. Notably, these languages have very poor isomorphism scores in relation to English. 

\paragraph{Why might graph-based methods work?} The goal for Procrustes is to find the ideal linear transformation $W_{ideal} \in \mathbb{R}^{\textbf{d} x \textbf{d}}$ to map the spaces, where \textbf{d} is the word embedding dimension. Seeds in Procrustes solve Equation \ref{eq:proc-eq} to find an approximation $W$ to $W_{ideal}$. Accordingly, the seeds can be thought of as samples from which one deduces the optimal linear transformation. This is a supervised learning problem, so when there are few seeds/samples, it is difficult to estimate $W_{ideal}$. Furthermore, the entire space $X$ is mapped by $W$ to a shared space with $Y$ meaning that \textit{every} point in $X$ is subject to a potentially inaccurate mapping $W$: the mapping extrapolates to the entire space. As graph-based methods, GOAT and SGM do not suffer this issue and can induce non-linear relationships. Graph methods can be thought of as a semi-supervised learning problem: even words that don’t serve as seeds are incorporated in the matching process. The graph manifold provides additional information that can be exploited.

Secondly, the dimension of the relationship between words in GOAT/SGM is much lower than for Procrustes. For the former, the relationship is one-dimensional: distance. As words for the Procrustes method are embedded in d-dimensional Euclidean space, their relationships have a magnitude \textit{and} a direction: they are $\{d + 1\}$-dimensional. It is possible that the lower dimension in GOAT/SGM makes them robust to noise, explaining why the graph-based methods outperform Procrustes in low-resource settings. This hypothesis should be investigated in follow-up studies.   

\section{Related Work}
\paragraph{BLI} Recent years have seen a proliferation of the BLI literature \cite[e.g.][]{ruder-etal-2018-discriminative,aldarmaki-etal-2018-unsupervised,joulin-etal-2018-loss,doval-etal-2018-improving,artetxe-etal-2019-bilingual,huang-etal-2019-hubless,patra-etal-2019-bilingual,zhang-etal-2020-overfitting,biesialska2020refinement}. Many use Procrustes-based solutions, which assume that embedding spaces are roughly isomorphic. \citet{wang-etal-2021-multi} argue that the mapping can only be piece-wise linear, and induce multiple mappings. 
\citet{ganesan-etal-2021-learning} learn an ``invertible neural network" as a non-linear mapping of spaces, and \citet{cao2018point} align spaces using point set registration. 
Many approaches address only high-resource languages. The tendency to evaluate on similar languages with high-quality data from similar domains hinders advancement in the field \cite{artetxe-etal-2020-call}.

\paragraph{BLI with OT}
Most similar to ours are BLI approaches which incorporate OT formulations using the Sinkhorn and/or Hungarian algorithms \cite[e.g.][]{alvarez-melis-jaakkola-2018-gromov,alaux2018unsupervised}. \citet{grave2019unsupervised} optimize ``Procrustes in Wasserstein Distance", iteratively updating a linear transformation and permutation matrix using Frank-Wolfe on samples from embedding spaces $\bf{X}$ and $\bf{Y}$. \citet{zhao-etal-2020-relaxed} and \citet{zhang-etal-2017-earth} also use an iterative procedure. \citet{ramirez2020novel} combine Procrustes and their Iterative Hungarian algorithms. 
\citet{xu-etal-2018-unsupervised-cross} use Sinkhorn distance in the loss function, and \cite{zhang-etal-2017-earth} use Sinkhorn to minimize 
distance between spaces.
\citet{haghighi-etal-2008-learning} use the Hungarian Algorithm for BLI from text. \citet{lianunsupervised} and \citet{alaux2018unsupervised} align all languages to a common space for multilingual BLI. The latter use Sinkhorn to approximate a permutation matrix in their formulation. \citet{zhao-etal-2020-semi} incorporate OT for semi-supervised BLI.

\section{Conclusion}
We perform bilingual lexicon induction from word embedding spaces of 40 language pairs, utilizing the newly-developed GOAT algorithm for graph-matching. 
Performance is strong across all pairs, especially on dissimilar languages with low-supervision. 
As the major use case of low-resource BLI and MT is dissimilar languages with low supervision, the strong performance of GOAT is an encouraging result for real-world applications.

\section*{Limitations}
Although we evaluate GOAT on 40 language pairs, this does not capture the full linguistic diversity of world languages. Languages of Eurasia are overrepresented, particularly the Indo-European family.
Each pair has at least one high-resource Indo-European language which uses the Latin script.
Future work should examine GOAT's performance when both languages are low-resource, and on an even broader diversity of languages from around the world.
Furthermore, graph-matching methods are also considerably slower than calculating the solution to the orthogonal Procustes problem on GPU, potentially limiting the former's usefulness when one must match large sets of words. 
Future work might examine the speed/accuracy trade-off between methods as embedding space size scales.

\section*{Acknowledgements}
This material is based on research sponsored by the Air Force Research Laboratory and Defense Advanced Research Projects Agency (DARPA) under agreement number FA8750-20-2-1001. The U.S. Government is authorized to reproduce and distribute reprints for Governmental purposes notwithstanding any copyright notation thereon. The views and conclusions contained herein are those of the authors and should not be interpreted as necessarily representing the official policies or endorsements, either expressed or implied, of the Air Force Research Laboratory and DARPA or the U.S. Government.  This material is based upon work supported by the United States Air Force under Contract No. FA8750‐19‐C‐0098.  Any opinions, findings, and conclusions or recommendations expressed in this material are those of the author(s) and do not necessarily reflect the views of the United States Air Force and DARPA.
 
\bibliography{anthology,custom}
\bibliographystyle{acl_natbib}

\clearpage
\appendix

\setcounter{table}{0}
\renewcommand{\thetable}{A\arabic{table}}

\section*{Appendix}

\begin{table}[htb]
\centering \footnotesize
\begin{tabular}{@{}l|rr|rr|r@{}}
\toprule
& \multicolumn{2}{c}{ *-to-en } & \multicolumn{2}{c}{ en-to-* } \\
& \underline{Full} & \underline{1-1} & \underline{Full} & \underline{1-1} & \underline{\# Embs} \\
\midrule
bn & 7588 & 4299 & 8467 & 4556 & 145350 \\
bs & 6164 & 4294 & 8153 & 4795 & 166505 \\
de & 10866 & 4451 & 14677 & 4903 & 200000 \\
en & n/a & n/a & n/a & n/a & 200000 \\
es & 8667 & 4445 & 11977 & 4866 & 200000 \\
et & 6509 & 4352 & 8261 & 4738 & 200000 \\
fa & 8510 & 4582 & 8869 & 4595 & 200000 \\
fr & 8270 & 4548 & 10872 & 4827 & 200000 \\
id & 9677 & 4563 & 9407 & 4573 & 200000 \\
it & 7364 & 4478 & 9657 & 4815 & 200000 \\
ja & 6819 & 4112 & 7135 & 4351 & 200000 \\
mk & 7197 & 4259 & 10075 & 4820 & 176947 \\
ms & 8140 & 4650 & 7394 & 4454 & 155629 \\
ru & 7452 & 4084 & 10887 & 4812 & 200000 \\
ta & 6850 & 4225 & 8091 & 4744 & 200000 \\
vi & 7251 & 4775 & 6353 & 4507 & 200000 \\
zh & 8891 & 4450 & 8728 & 4381 & 200000 \\
\bottomrule
\end{tabular}
\caption{Size of train/test sets before (Full) \& after making one-to-one (1-1), with \# of embeddings used.}
\label{tab:datasizes}
\end{table}

\begin{table}[htb]
\centering \footnotesize
    \begin{tabular}{@{}r|rr|rr@{}}
    \toprule
    & \multicolumn{2}{c}{\underline{ en-de }} & \multicolumn{2}{c}{\underline{ ru-en }} \\
    \midrule
    Seeds & Rand. & Bary. & Rand. & Bary.   \\ 
    100   &   45.7	& 45.9  &   49.6    &	50.4 \\
    200   &   47.4	& 47.5  &   52.5    &	52.5 \\
    500   &   52.3	& 51.9  &   55.4    &	55.6 \\
    1000  &   54.6	& 54.9  &   58.3    &	58.1 \\
    2000  &   61.5	& 61.4  &   67.1    &	67.1 \\
    4000  &   74.2	& 74.2  &   89.3    &	89.3 \\
    \bottomrule
    \end{tabular}
    \caption{SGM with barycenter vs. randomized initialization for languages used in \citet{marchisio-etal-2021-analysis-euclidean}. The difference is negligible.}
\end{table}

\paragraph{Unsupervised Performance}
For some highly-related languages, GOAT performs well even with no seeds (unsupervised). GOAT scores 48.8 on English$\rightarrow$German, 34.5 on German$\rightarrow$English, 62.4 on English$\rightarrow$Spanish, and 19.6 on Spanish$\rightarrow$English with no supervision. Particularly striking is the unsupervised performance on highly-related languages: $>$87 on Italian$\leftrightarrow$French and $>$90 for Spanish$\leftrightarrow$Portuguese.  We suspect that that the word embedding spaces are highly isomorphic for these language pairs, allowing GOAT (and sometimes SGM) to easily recover the translations.

\paragraph{Iterative} Results of Iterative Procrustes (IterProc), Iterative SGM (IterSGM), and Iterative GOAT (IterGOAT) are in Table \ref{tab:iterative}.  
We run the Iterative Procrustes and Iterative SGM procedures of \citet{marchisio-etal-2021-analysis-euclidean} with stochastic-add.  Here, Procrustes [or SGM] is run in source$\leftrightarrow$target directions, hypotheses are intersected, and H random hypotheses are added to the gold seeds and fed into subsequent runs of Procrustes [SGM].  The next iteration adds 2H hypotheses, repeating until all hypotheses are chosen. We set $H = 100$ and create an analogous iterative algorithm for GOAT, which we call Iterative GOAT.

IterSGM/GOAT perform similarly across conditions, with a few exceptions where either performs very strongly with no supervision: IterGOAT scores 49.2, 45.2, 34.4, 58.2, and 55.9 for En-De, En-Fa, De-En, Id-En, and Ru-En, respectively, and IterSGM scores 57.3 for En-Ru. On Chinese$\leftrightarrow$English, IterGOAT underperforms IterSGM, similar to GOAT's underperformance of SGM in the single run.

Similar to \citet{marchisio-etal-2021-analysis-euclidean}, we find that IterProc compensates for an initial poor first run and outperforms IterSGM with a moderate amount of seeds (100+). Extending to the very lowest seeds sizes (0-75), however, IterSGM/IterGOAT are superior. With 25 seeds, IterProc fails for all language pairs except En$\leftrightarrow$De and En$\leftrightarrow$Ru, scoring $P@1 < 5$. IterSGM and IterGOAT, however, perform reasonably well for most language pairs with 25 seeds, suggesting that the graph-based framing is the better approach for low-seed levels. At the highest supervision level (2000+ seeds), IterSGM/IterGOAT again tends to be superior.
 
Differences are minor btwn using GOAT or SGM in the iterative or system combination experiments. This results suggests that mixing Procrustes and graph-based framings is helpful for BLI, regardless of which algorithm one picks. It is interesting to contemplate what other problems might benefit from examination from multiple mathematical framings in one solution, as each may have complementary benefits.  

\onecolumn
\begin{sidewaystable}
\centering \setlength\tabcolsep{1pt} \footnotesize
\begin{tabular}{@{}l|cccc|cccc|cccc|cccc|cccc|cccc|cccc|cccc@{}} 
\hline
 & \multicolumn{4}{c}{\underline{en-bn}} & \multicolumn{4}{c}{\underline{en-bs}}  & \multicolumn{4}{c}{\underline{en-et}} & \multicolumn{4}{c}{en-fa}  & \multicolumn{4}{c}{\underline{en-id}} & \multicolumn{4}{c}{\underline{en-mk}} & \multicolumn{4}{c}{\underline{en-ms}} &  \multicolumn{4}{c}{\underline{en-ta}}  \\
 \hline
\multicolumn{1}{l}{\textit{Seeds}} & \textit{P} & \textit{S} &\textit{G} & $\Delta$ & \textit{P} & \textit{S} &\textit{G} & $\Delta$ & \textit{P} & \textit{S} &\textit{G} & $\Delta$ & \textit{P} & \textit{S} &\textit{G} & $\Delta$ & \textit{P} & \textit{S} &\textit{G} & $\Delta$ & \textit{P} & \textit{S} &\textit{G} & $\Delta$ & \textit{P} & \textit{S} &\textit{G} & $\Delta$ & \textit{P} & \textit{S} & \textit{G} & $\Delta$ \\
0 & 0.0 & 0.0 & \textbf{0.1} & \textcolor{darkgreen}{\textit{(+0.1)}} & 0.0 & 0.0 & 0.0 & \textcolor{gray}{\textit{(+0.0)}} & 0.0 & 0.0 & \textbf{0.2} & \textcolor{darkgreen}{\textit{(+0.2)}} & 0.0 & 0.2 & \textbf{0.5} & \textcolor{darkgreen}{\textit{(+0.3)}} & 0.0 & 0.2 & \textbf{2.9} & \textcolor{darkgreen}{\textit{(+2.7)}} & 0.1 & 0.0 & \textbf{0.1} & \textcolor{darkgreen}{\textit{(+0.1)}} & 0.0 & 0.0 & \textbf{0.1} & \textcolor{darkgreen}{\textit{(+0.1)}} & \multicolumn{1}{r}{0.0} & \multicolumn{1}{r}{0.0} & \multicolumn{1}{r}{0.0} & \multicolumn{1}{r}{\textcolor{gray}{\textit{(+0.0)}}} \\
\hline

\end{tabular}

\caption{Full Results (1 of 2): P@1 of Procrustes (P), SGM (S) or GOAT (G). $\Delta$ is gain/loss of GOAT over SGM.}
\label{tab:single-full1}
\end{sidewaystable}
\twocolumn

\begin{table*}[htb]
    \centering \setlength\tabcolsep{4pt} \footnotesize
    \begin{tabular}{l|rrrr|rrrr|rrrr|rrrr}
\toprule
 & \multicolumn{4}{c}{\underline{vi-en}} & \multicolumn{4}{c}{\underline{zh-en}} & \multicolumn{4}{c}{\underline{ru-en}} & \multicolumn{4}{c}{\underline{de-en}} \\
\hline 
\multicolumn{1}{l}{\textit{Seeds}} & \textit{P} & \textit{S} & \textit{G} & $\Delta$ & \textit{P} & \textit{S} & \textit{G} & $\Delta$ & \textit{P} & \textit{S} & \textit{G} & $\Delta$ & \textit{P} & \textit{S} & \textit{G} & $\Delta$ \\
0 & 0.0 & 0.0 & 0.1 & \textcolor{darkgreen}{\textit{(+0.1)}} & 0.0 & 0.0 & 0.0 & \textcolor{gray}{\textit{(+0.0)}} & 0.0 & 0.4 & \textbf{2.8} & \textcolor{darkgreen}{\textit{(+2.4)}} & 0.0 & 0.6 & \textbf{34.5} & \textcolor{darkgreen}{\textit{(+33.9)}} \\
25 & 0.1 & 1.0 & \textbf{3.3} & \textcolor{darkgreen}{\textit{(+2.3)}} & 0.1 & 6.8 & \textbf{9.3} & \textcolor{darkgreen}{\textit{(+2.5)}} & 0.3 & 49.1 & \textbf{54.4} & \textcolor{darkgreen}{\textit{(+5.3)}} & 0.2 & 30.3 & \textbf{34.1} & \textcolor{darkgreen}{\textit{(+3.8)}} \\
50 & 0.2 & 27.4 & \textbf{41.7} & \textcolor{darkgreen}{\textit{(+14.3)}} & 0.6 & \textbf{13.1} & 12.5 & \textcolor{red}{\textit{(-0.6)}} & 1.0 & 49.5 & \textbf{54.8} & \textcolor{darkgreen}{\textit{(+5.3)}} & 0.6 & 42.5 & \textbf{47.6} & \textcolor{darkgreen}{\textit{(+5.1)}} \\
75 & 0.3 & 29.8 & \textbf{45.2} & \textcolor{darkgreen}{\textit{(+15.4)}} & 1.3 & 12.6 & \textbf{14.1} & \textcolor{darkgreen}{\textit{(+1.5)}} & 2.1 & 49.9 & \textbf{54.8} & \textcolor{darkgreen}{\textit{(+4.9)}} & 1.5 & 42.6 & \textbf{47.0} & \textcolor{darkgreen}{\textit{(+4.4)}} \\
100 & 0.5 & 34.4 & \textbf{47.1} & \textcolor{darkgreen}{\textit{(+12.7)}} & 2.4 & \textbf{15.2} & 14.9 & \textcolor{red}{\textit{(-0.3)}} & 4.1 & 50.4 & \textbf{55.1} & \textcolor{darkgreen}{\textit{(+4.7)}} & 3.0 & 41.9 & \textbf{47.6} & \textcolor{darkgreen}{\textit{(+5.7)}} \\
\bottomrule
    \end{tabular}
\caption{Full Results (2 of 2): P@1 of Procrustes (P), SGM (S) or GOAT (G). $\Delta$ is gain/loss of GOAT over SGM.}
\label{tab:single-full2}
\end{table*}


\begin{table*}[hbt]
    \centering \setlength\tabcolsep{3.5pt} \footnotesize
    \begin{tabular}{l|rrr|rrr|rrr|rrr|rrr|rrr}
    \toprule
        & \textit{IP} & \textit{IS} & \textit{IG} & \textit{IP} & \textit{IS} & \textit{IG} & \textit{IP} & \textit{IS} & \textit{IG} & \textit{IP} & \textit{IS} & \textit{IG} & \textit{IP} & \textit{IS} & \textit{IG} & \textit{IP} & \textit{IS} & \textit{IG} \\
        \hline
        \textit{Seeds} & \multicolumn{3}{c}{\underline{en-bn}} &  \multicolumn{3}{c}{\underline{bn-en}} &  \multicolumn{3}{c}{\underline{en-bs}} &  \multicolumn{3}{c}{\underline{bs-en}} &  \multicolumn{3}{c}{\underline{en-de}} &  \multicolumn{3}{c}{\underline{de-en}}  \\
        0 & 0.0 & 0.0 & 0.0 & 0.0 & 0.0 & 0.1 & 0.0 & 0.0 & 0.0 & 0.0 & 0.1 & 0.1 & 0.0 & 0.3 & \textbf{49.2} & 0.1 & 0.3 & \textbf{34.4} \\
        25 & 0.6 & \textbf{37.3} & \textit{35.6} & 3.2 & \textbf{47.3} & \textit{46.8} & 0.2 & \textbf{52.3} & \textit{51.2} & 0.1 & \textbf{56.5} & \textit{56.1} & \textbf{61.5} & 49.8 & 49.7 & \textbf{58.8} & 47.6 & 47.2 \\
        50 & 1.6 & \textbf{37.8} & \textit{35.8} & 24.3 & \textbf{47.5} & \textit{47.2} & 48.3 & \textbf{52.2} & \textit{51.4} & 4.2 & \textbf{57.2} & \textit{56.2} & \textbf{61.5} & 49.9 & 49.6 & \textbf{59.0} & 47.8 & 47.9 \\
        75 & \textbf{38.8} & 37.3 & 36.7 & \textbf{51.9} & 48.3 & 47.6 & \textbf{54.7} & 52.3 & 51.5 & \textbf{58.4} & 57.1 & 56.4 & \textbf{61.5} & 50.4 & 49.9 & \textbf{59.0} & 48.5 & 47.9 \\
        100 & \textbf{44.4} & 37.7 & 36.5 & \textbf{53.2} & 48.2 & 47.9 & \textbf{54.6} & 52.5 & 51.1 & \textbf{59.0} & 57.1 & 56.8 & \textbf{62.1} & 50.2 & 50.0 & \textbf{59.4} & 48.4 & 48.1 \\
        200 & \textbf{43.3} & 38.9 & 37.2 & \textbf{52.1} & 48.0 & 48.1 & \textbf{55.1} & 52.9 & 52.4 & \textbf{59.3} & 58.8 & 57.7 & \textbf{62.0} & 50.9 & 50.4 & \textbf{59.7} & 49.3 & 48.8 \\
\bottomrule
    \end{tabular}
    \caption{P@1 of Iterative Procrustes (IP), Iterative SGM (IS), and Iterative GOAT (IG). \textbf{Highest per row}. IterSGM/IterGOAT are \textit{italicized} when outperforming IterProc but are not highest in row.}
    \label{tab:iterative} 
\end{table*}

\begin{table*}[htb]
    \centering \setlength\tabcolsep{2.1pt} \small
    \begin{tabular}{l@{}|l@{\hspace{-1mm}}>{\color[gray]{0.2}}rrrrr@{\hspace{2mm}}|>{\color[gray]{0.2}}rrrrr||l@{\hspace{-1mm}}>{\color[gray]{0.2}}rrrrr@{\hspace{2mm}}|>{\color[gray]{0.2}}rrrrr@{}}
        & &  \multicolumn{5}{c}{\underline{en-to-*}} & \multicolumn{5}{c}{\underline{*-to-en}} & & \multicolumn{5}{c}{\underline{en-to-*}} & \multicolumn{5}{c}{\underline{*-to-en}} \\
        & &  & \multicolumn{2}{c}{SGM} & \multicolumn{2}{c}{GOAT} &  & \multicolumn{2}{c}{SGM} & \multicolumn{2}{c}{GOAT} & &  &  \multicolumn{2}{c}{SGM} & \multicolumn{2}{c}{GOAT} &  & \multicolumn{2}{c}{SGM} & \multicolumn{2}{c}{GOAT} \\
        Seeds & & \textit{Prev} &\textit{-PP} &\textit{-PS} &\textit{-PP} &\textit{-PG} & \textit{Prev} &\textit{-PP} &\textit{-PS} &\textit{-PP} &\textit{-PG} & & \textit{Prev} &\textit{-PP} &\textit{-PS} &\textit{-PP} &\textit{-PG} & \textit{Prev} &\textit{-PP} &\textit{-PS} &\textit{-PP} &\textit{-PG} \\
        \hline
         & bn &  &  &  &  &  &  &  &  &  &  & mk &  &  &  &  &  &  &  &  &  & \\
        0 &  & \textit{0.1} & 0.0 & 0.1 & 0.0 & 0.1 & \textit{0.1} & 0.0 & 0.0 & 0.1 & 0.0 &  & \textit{0.1} & 0.0 & 0.1 & 0.0 & 0.0 & \textit{0.1} & 0.1 & 0.0 & 0.0 & 0.1 \\
        25 &  & \textit{37.3} & 44.0 & 0.5 & \textbf{44.3} & 0.5 & \textit{47.3} & 52.8 & 8.4 & \textbf{53.1} & 6.2 &  & \textit{59.4} & \textbf{64.4} & 1.2 & 63.9 & 8.0 & \textit{65.2} & 68.7 & 1.1 & \textbf{68.8} & 0.6 \\
        50 &  & \textit{37.8} & 43.1 & 2.0 & \textbf{44.3} & 3.1 & \textit{47.5} & 52.1 & 14.4 & \textbf{53.0} & 48.2 &  & \textit{62.1} & \textbf{64.0} & 63.6 & 63.9 & 62.8 & \textit{66.3} & 68.4 & 68.1 & \textbf{68.5} & 67.8 \\
        75 &  & \textit{38.8} & \textbf{44.7} & 38.8 & \textbf{44.7} & 39.5 & \textit{51.9} & 53.2 & 49.2 & \textbf{53.5} & 49.1 &  & \textit{61.7} & 63.8 & 63.7 & \textbf{64.3} & 63.3 & \textit{66.8} & \textbf{69.2} & 67.3 & 68.8 & 67.9 \\
\hline
    \end{tabular}
    \caption{Full Results: P@1 of Combination Experiments. SGM-PP starts with SGM, ends with Procrustes. SGM-PS: IterProc then SGM. GOAT-PP: start GOAT, end Proc. GOAT-PG: IterProc then GOAT. Previous best of all other experiments is in the \textit{Prev} column. \textit{Prev} here includes iterative results from Table \ref{tab:iterative}.} 
    \label{tab:combined}
\end{table*}

\end{document}